\documentclass[12pt]{article}
\usepackage{amssymb,amsmath,amsthm}

\usepackage{graphicx}
\usepackage{natbib}
\usepackage{url} 

\newcommand{\blind}{0}

\usepackage{wasysym}
\usepackage{hyperref}
\usepackage{algorithm}
\usepackage{algorithmic}

\usepackage[abs]{overpic}

\newcommand{\response}{y}

\newcommand{\predictor}{x}
\newcommand{\predictors}{\boldsymbol{x}}

\newcommand{\estf}{f}

\newcommand{\cond}{C}
\newcommand{\sect}{S}
\newcommand{\threshold}{\sigma}

\newcommand{\predictorvals}{\boldsymbol{u}}

\newcommand{\bu}{\boldsymbol{u}}
\newcommand{\bv}{\boldsymbol{v}}

 \usepackage{array}

\usepackage{listings}

\newcommand{\pkg}[1]{{\normalfont\fontseries{b}\selectfont #1}}

\let\code=\texttt

\begin{document}

\def\spacingset#1{\renewcommand{\baselinestretch}%
{#1}\small\normalsize} \spacingset{1}

\if0\blind
{
  \title{\bf Interactive slice visualization for exploring machine learning models}
  \author{Catherine B. Hurley,  Mark O'Connell, Katarina Domijan\\
    \hspace{.2cm}\\
    Department of Mathematics and Statistics, Maynooth University}
  \maketitle
} \fi

\if1\blind
{
  \bigskip
  \bigskip
  \bigskip
  \begin{center}
    {\LARGE\bf Interactive slice visualization for exploring machine learning models}
\end{center}
  \medskip
} \fi

\bigskip
\begin{abstract}
Machine learning models fit complex algorithms to arbitrarily large datasets. 
These algorithms are well-known to be high on performance and low on interpretability. We use interactive visualization of slices
of predictor space to address the
 interpretability deficit; in effect opening up the black-box of machine learning algorithms, for the purpose of interrogating, explaining, validating and comparing model fits.
 Slices are specified directly through interaction, or using various touring algorithms designed to visit high-occupancy sections, or regions where
 the model fits have interesting  properties.
The methods presented here are implemented in the R package \pkg{condvis2}.
\end{abstract}

\noindent%
\vfill

\newpage
\spacingset{1.5} 

\section{Introduction}
Machine learning models fit complex algorithms to 
extract predictions from datasets.
Numerical model summaries such as mean squared residuals and feature importance measures are
commonly used for assessing model performance, feature importance and for comparing various fits.
Visualization is a powerful way of drilling down, going beyond numerical summaries to explore how predictors 
impact on the fit, assess goodness of fit and compare multiple fits in different regions of predictor space,
 and perhaps ultimately developing improved fits. Coupled with interaction, visualization becomes an even more
 powerful model exploratory tool.

Currently, explainable artificial intelligence (XAI) is a very active research topic, with the goal of making models
understandable to humans.
There have been many efforts  to use visualization to understand machine learning fits in a model-agnostic way.
Many of these show how features locally explain a fit \citep{lime,lundberg}.
 \cite{staniak} give an overview of R packages for local explanations and present some nice visualizations.
 Other visualizations such as partial dependence  plots \citep{pdp} shows how a predictor affects the fit on average.
Drilling down, more detail is obtained by exploring the effect of a designated predictor on the fit, conditioning on  fixed values of
other predictors, for example using the individual conditional expectation (ICE) curves of \cite{ice}.
Interactive visualizations are perhaps under-utilized in this context.
\cite{grammar} offer a recent discussion. \cite{vine} uses small multiple displays of clustered ICE curves
in an interactive framework to visualize interaction effects.

Visualizing data via conditioning or slicing was popularised by the  ``small multiples'' of Tufte \citep{tufte}
and the trellis displays of \cite{trellis}.
Nowadays, the concept is widely known as {\it faceting}, courtesy of \cite{ggplot2}.
\cite{Wilkinson2005} (chapter 11) gives a comprehensive description.
In the context of machine learning models, the conditioning concept is used in
ICE plots, which show a family of curves giving the fitted response
for one predictor, fixing other predictors at observed values.
 These ICE plots simultaneously show all observations
and overlaid fitted curves, one for each observation in the dataset.
Partial dependence plots  which show the average of the ice curves are more popular
but these are known to suffer from bias in the presence of correlated predictors.
A recent paper  \citep{wire} gives a comparison of these and other model visualization techniques
based on conditioning.

Visualization along with interactivity is a natural and powerful way of exploring data;
so-called {\it brushing} \citep{wxs} is probably the best-known example.
Other data visualization applications have used interaction in creative ways,
for high-dimensional data \pkg{ggobi} (see for example \cite{ggobi}) offers various kinds of  low-dimensional dynamic projection tours
while the recent R package  \pkg{loon}  \citep{loon} has a graph-based interface for moving through series of scatterplots.
The interactive display paradigm has also been applied to exploratory modelling analysis, for example \cite{klimt} describes an application for exploratory analysis of trees.
With interactive displays, the data analyst has the ability to sift through many plots quickly and easily, discovering 
interesting and perhaps unexpected patterns.

In this paper, we present model visualization techniques  based on slicing high-dimensional space, where interaction is used to navigate the slices through the space.
The idea of using interactive visualization in this way was introduced in \cite{condvis-jss}.
The basic concept is to fix the values of all but one or two predictors, and to display the conditional fitted curve or surface.
Observations from a slice close to the fixed predictors are overlaid on the curve or surface.
The resulting visualizations will  show how predictors affect
the fitted model and the model goodness of fit, and how this varies as the slice is navigated through predictor space.
We also describe touring algorithms for exploring predictor space. These algorithms make
 conditional visualization a practical and valuable tool for model exploration as dimensions increase. 
 Our techniques are model agnostic and are appropriate for any regression or classification problem.
 The concepts of conditional visualization are also relevant for  ``fits'' provided by clustering and density estimation algorithms.
  Our model visualization techniques are implemented in our   R package   \pkg{condvis2} \citep{condvisP2}, 
 which provides a highly-interactive application for model exploration.

The outline of the paper is as follows.
 In Section 2 we describe the basic ideas of conditional visualization for model fits, and follow that with 
our tour constructions for visiting interesting and relevant slices of data space.
Section 3 focuses on our  implementation,  and describes the embedding of conditional model visualizations in an interactive application.
In Section 4  we present  examples, illustrating how our methods
are used to understand predictor effects,  explore lack of fit and to compare multiple fits.
We conclude with a discussion.

\section{Slice visualization and construction }

In this section we describe the construction of slice visualizations for exploring  machine learning models. We
begin with notation and terminology. Then we explain how observations near a slice are identified, and then visualized
using a color gradient.
We present new touring algorithms designed to visit high-occupancy slices and slices where
 model fits have interesting  properties. In practical applications, these touring algorithms mean our model exploration techniques
 are useful for exploring fits with up to 30 predictors.

Consider
 data $\{\predictors_i,\response_i\}_{i=1}^{n}$, where $\predictors_i=
(\predictor_{i1},...,\predictor_{ip})$ is a vector of predictors and
$\response_i$ is the response. Let  $\estf$ denote a fitted model  that maps the
predictors $\predictors$ to fitted responses $\estf(\predictors)$.  
(In many applications we will have two or more fits which we wish to compare, but we use just one here for ease of explanation.)
Suppose  there are just a few predictors of primary interest. We call these the \emph{section}  predictors and index them by $\sect$.
The remaining predictors are called  \emph{conditioning}  predictors, indexed by $\cond$.
Corresponding to $\sect$ and $\cond$, partition the feature coordinates $\predictors$ into
$\predictors_\sect$ and $\predictors_\cond$. 
Similarly, let $\predictors_{i\sect}$ and $\predictors_{i\cond}$
denote the coordinates of observation $i$ for the predictors in $\sect$ and $\cond$ respectively.
We have
interest in observing the relationship between the response $\response$, fit  $\estf$, 
and $\predictors_\sect$, conditional on
$\predictors_\cond$. 
For our purposes, a section or slice is constructed as a region around a single point
 in the space of $\cond$, i.e. $\predictors_\cond = \predictorvals_\cond$,
where $ \predictorvals_\cond$ is called the \emph{section point}.

\subsection{Visualizations }

Two related visualizations show the fit and the data.
This first display is the so-called  \emph{section plot} which shows how the fit $\estf$ varies over the
predictors in $\predictors_\sect$. The second display shows plots of the predictors in $\predictors_\cond$
and the current setting of the section point $ \predictorvals_\cond$.
We call these the \emph{condition selector plots}, as the section point $ \predictorvals_\cond$ is under interactive control.

More specifically, the section plot consists of
$\estf(\predictors_\sect, \predictors_\cond = \predictorvals_\cond)$ versus ${\predictors_\sect}$,
shown on a grid covering $\predictors_\sect$,
overlaid on a subset of observations $(\predictors_{i\sect}, y_i)$, where $\predictors_{i\cond}$
is near the designated section point  $\predictorvals_\cond$. 
For displaying model fits, we use $|\sect| = 1,2$, though having more  variables in $S$ would be possible with faceted displays.

\subsubsection{Similarity scores and color}

A key feature of the section plot is that only observations local to the section point $\predictorvals_\cond$ are included.
To determine these local observations, 
we start with a distance measure $d$,  and for each observation, $i=1,2,\ldots, n$, we compute how far it is from the section point $\predictorvals_\cond$ as
\begin{align}
  d_i = d(\predictorvals_\cond, \predictors_{i\cond}).
\end{align}
This distance is converted to a similarity score as 
 \begin{align}
\label{eq:simweight}
  s_i = \max \left( 0, 1 - \dfrac{d_i}{\threshold}\right)
\end{align}
where $\threshold > 0$ is a threshold parameter. Distances exceeding the threshold $\threshold$ are accorded a
similarity score of zero. Points on the section, that is, identical to the section point $\predictorvals_\cond$,
receive the maximum similarity of 1.
Plotting colors for points are then faded to the background white color
using these similarity scores. 
Points with a similarity score of zero become white, that is, are not shown.
Non-zero similarities are binned into  equal-width intervals. 
The colors of observations whose similarity belongs to the right-most interval are left unchanged.
Other observations are faded to white, with the amount of fade decreasing from the first interval to the last.

\subsubsection{Distances for similarity scores}

We use two different notions of ``distance'' in calculating similarity scores.
The first is a Minkowski distance between numeric coordinates (Equation \ref{minkowski}).
For two vectors $\bu$ and $\bv$, where $C_{num}$ indexes  numeric predictors and its complement $C_{cat}$ indexes the categorical predictors
in the conditioning set $C$,
\begin{align}
\label{minkowski}
   d_M(\bu,\bv) =  \left\{
                \begin{array}{ll}
                  \left(\sum_{j \in C_{num}} | \bu_j - \bv_j|^q \right)^{1/q}  & \text{ if } \bu_k = \bv_k \; \forall k \in C_{cat}\\
                  \infty & \text{otherwise.}
                  \end{array} \right.
\end{align}
 In practice we use   Euclidean distance given by  $q=2$ and the maxnorm distance
 which is the limit as $q \rightarrow \infty$ (equivalently $\max_j  | \bu_j - \bv_j|$).
 With the Minkowski distance, points whose categorical coordinates do not match those of the section 
 $\predictorvals_\cond$ exactly will receive a similarity of zero and will not be visible in the section plots.
 Using Euclidean distance, visible observations in the section plot will be in the  hypersphere of radius $\sigma$
 centered at $\predictorvals_\cond$. Switching to the maxnorm distance means that visible observations will be in the unit hypercube
 with sides of length $2\sigma$.

If there are many categorical conditioning predictors, requiring an exact match on categorical predictors could
mean that there are no visible observations.
For this situation, we  include a Gower distance \citep{gower} 
given in Equation \ref{gower}
which combines absolute differences in numeric coordinates and mismatch counts in
categorical coordinates,
\begin{align}
\label{gower}
d_G(\bu,\bv) = \sum_{k \in C_{num} }\frac{| u_k - v_k|}{R_k} +  \sum_{k \in C_{cat}}   1 \left [ u_k  \neq  v_k  \right] 
\end{align}
where $R_k$ is the range of the $k$th predictor in $C_{num}$.

\subsubsection{A toy example }
\begin{figure}[htbp]
\begin{center}
\begin{tabular}{ccc}
 \includegraphics[scale=.6,trim={0 60 0 0},clip]{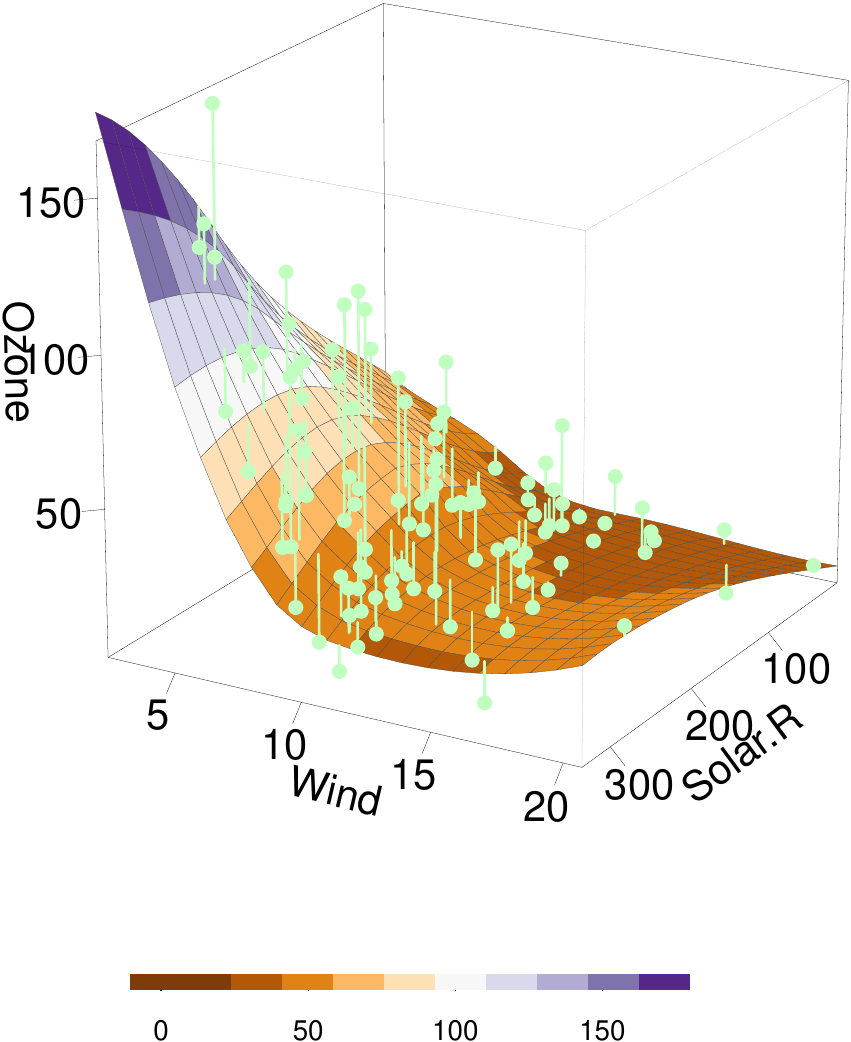} & \includegraphics[scale=.4]{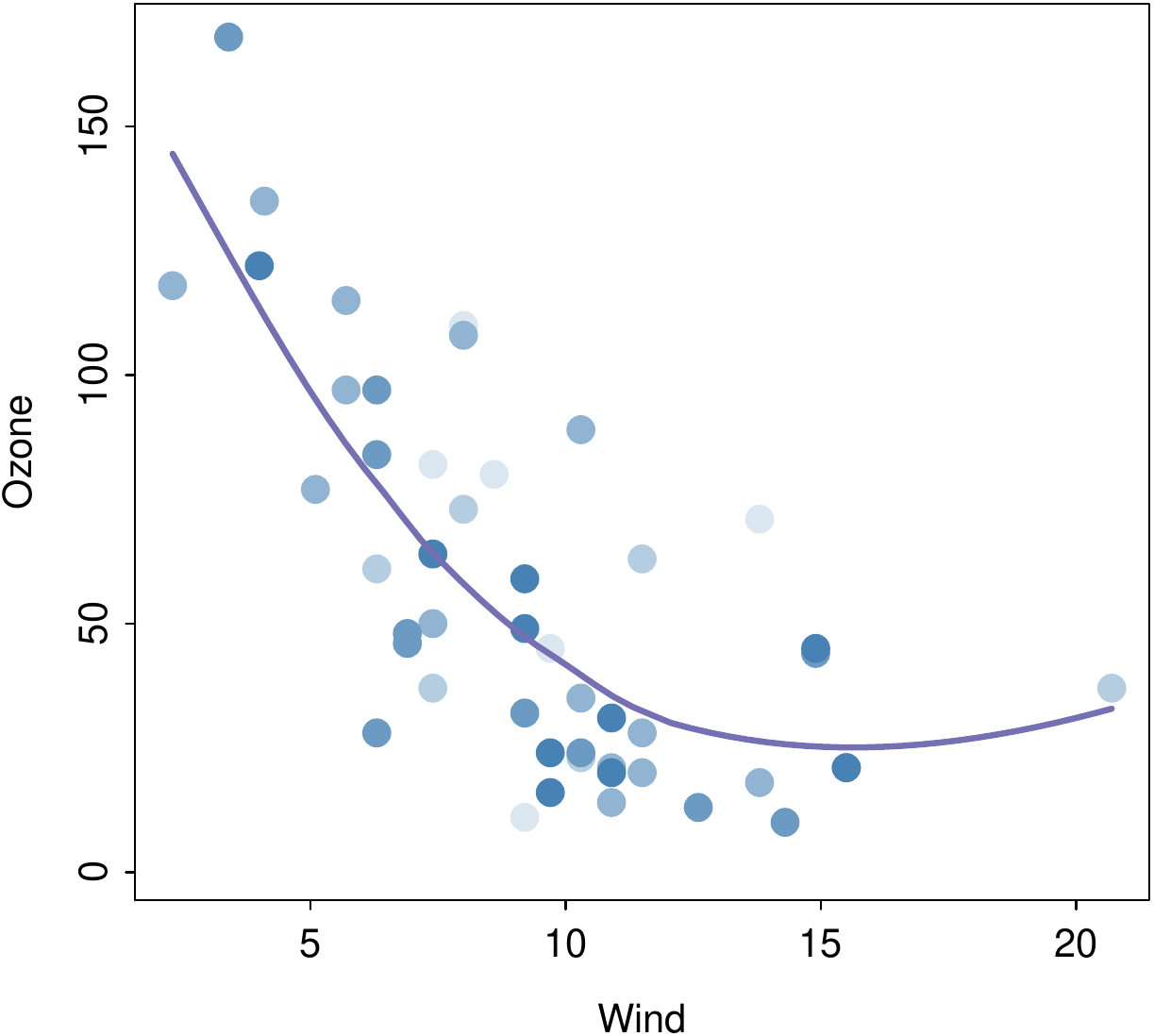} &
 \includegraphics[scale=.4]{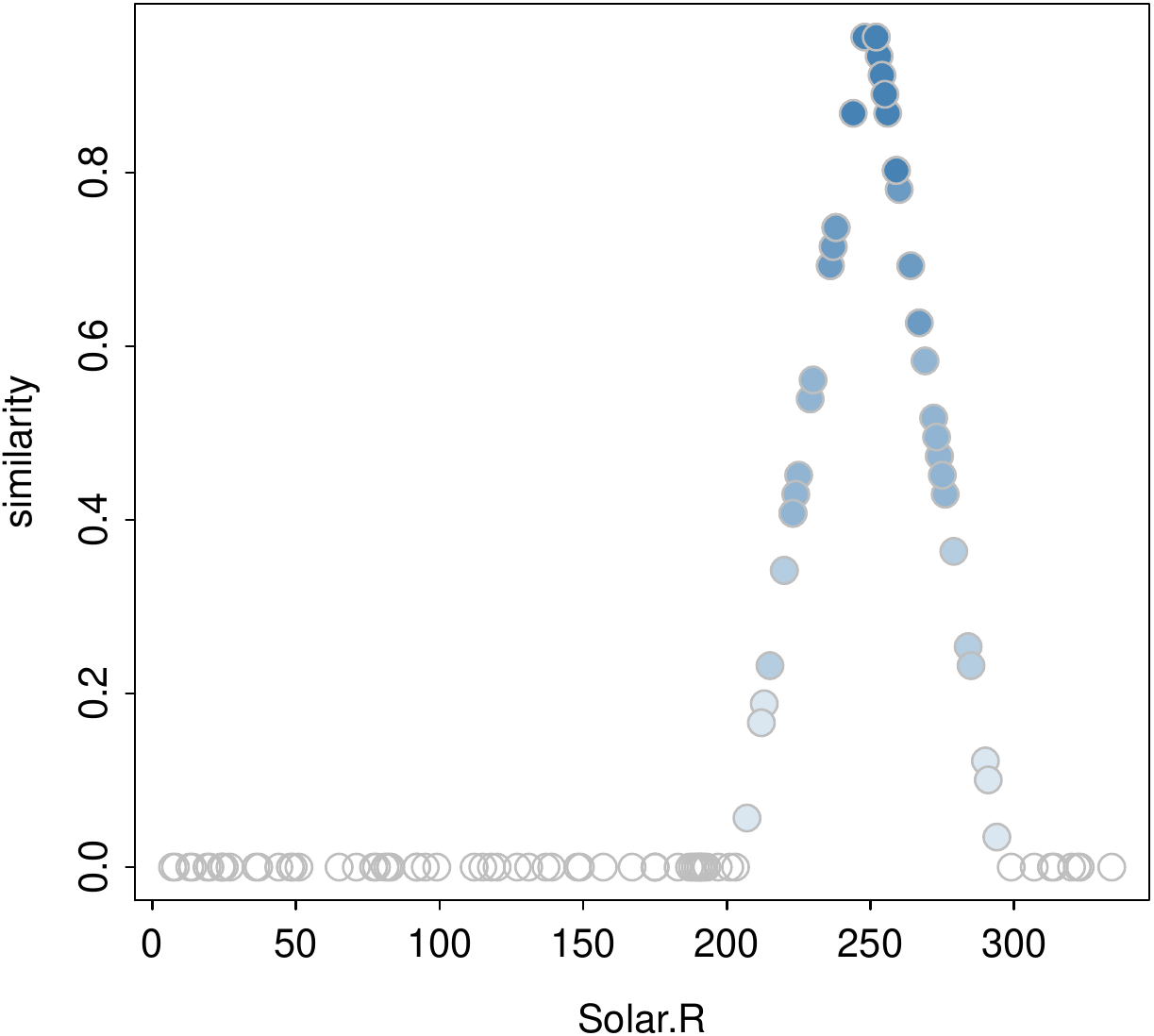} \\
 (a) 3d-surface & (b) 2d-slice at Solar.R = 250 &(c) Fading point color
 \end{tabular}
\caption{Illustration of relationship between distance and color in section plot, using the three variable ozone data: (a) data and loess fit as a surface, (b) a section plot showing fit versus Wind conditioning on Solar.R=250, (c) distance  to  Solar.R=250  represented by color.
}
\label{ozone}
\end{center}
\end{figure}

To demonstrate the ideas of the previous subsections, we use an  illustration in the simple setting with just two predictors.
Figure \ref{ozone}(a)
shows a loess surface relating Ozone to Solar.R and Wind in the  air quality data \citep{gmda}.
Consider $S$ = Wind and $C$ = Solar.R, and fix the value of Solar.R as $\predictorvals_\cond = 250$.
Figure \ref{ozone}(b) is the resulting section plot, where we see how the surface varies with Wind, with Solar.R fixed at 250. 
Only observations
with Solar.R near 250 (here about  250 $\pm$ 50) are shown.
 Observations in this window receive a similarity
score related to their distance from 250, which is used to fade the color by distance. 
Observations outside the window receive a similarity
score of zero and are not displayed in Figure \ref{ozone}(b).  In Figure \ref{ozone}(c), the similarity scores assigned to observations using their distance to Solar.R=250 are plotted, non-zero
similarity scores are faded by decreasing similarity.

From the section plot in Figure \ref{ozone}(b) it is apparent that there is just one observation at Wind $\approx$ 20,
so the fit in this region may not be too reliable.
By decreasing the Solar.R value to $\predictorvals_\cond = 150$ and then to 
50  we learn that the dependence of Ozone on Wind also decreases.

\subsection{Choosing section points }

The simplest way of specifying $ \predictorvals_\cond$ is to choose a particular observation, or to  supply  a value of each predictor in $C$.
As an alternative to this, we can find areas where the data lives and visualize these. 
This is particularly important as the number of predictors increases: the well-known
\emph{curse of dimensionality} \cite{Bellman} implies that as the dimension of the conditioning space
increases, conditioning on arbitrary predictor settings will yield mostly empty sections.
Or, we can look for interesting sections
exhibiting features such as lack of fit, curvature or interaction. In the case of multiple fits, we can chase differences between them.

We describe algorithms for the construction of \emph{tours}, which for our purposes are a series of section points  $\{ \predictorvals^k_\cond, k=1,2, \ldots, l \}$.
The tours are visualized by section plots $\estf(\predictors_\sect = {\predictors_\sect}^g, \predictors_\cond = \predictorvals^{k}_\cond)$,
showing slices formed around the series of section points.
We note that  the tours presented here are quite different to  grand tours \citep{asimov} and guided tours \citep{guidedtour}, which are formed
as sequences of projection planes and do not involve slicing.

\subsubsection{Tour construction: visiting regions with data }

The simplest strategy to find where the data lives  is to pick random observations and use their coordinates 
for the conditioning predictors as sections points.
We call this the randomPath tour. 
Other touring options cluster the data using the variables in $C$, and use the cluster centers as section points.
It is important to note that we are not trying to identify actual clusters in the data, rather to
visit the parts of $C$-predictor space where observations are located. 
We consider two tours based on clustering algorithms:
(i) kmeansPath which uses centroids of k-means clusters as sections
and (ii) kmedPath which uses medoids of k-medoid clustering, available from the 
pam algorithm of package \pkg{cluster} \citep{cluster}. Recall that medoids are observations in the dataset, so slices around them are
guaranteed to have at least one observation.

Both kmeansPath and kmedPath work for categorical as well as numerical variables.
kmeansPath standardizes numeric variables and hot-encodes categorical variables. kmedPath uses 
a distance matrix based on standardized Euclidean distances for numeric variables and the Gower distance \citep{gower} for variables of mixed type, as provided by
\code{daisy}
from package \pkg{cluster}.  For our application we are not concerned
with optimal clustering or choice of number of clusters, our goal is simply to visit regions where the data live.

To evaluate our tour algorithms, we calculate
randomPath, kmeansPath and kmedPath tours of length $l=30$ on datasets of 2,000 rows and 15 numeric variables
obtained from the Ames \citep{ames} and  Decathlon  \citep{GDAdata} datasets.  For comparison, we also use simulated independent Normal and Uniform datasets of the same dimension.
 The results  are  summarized  Table \ref{amestab}.
\begin{table}[ht]
\caption{Average number of visible observations in ($\sigma$=1) maxnorm slices at 30 section points and in parentheses their total
similarity   selected with randomPath, kmeansPath and kmedPath
from Decathlon and Ames   datasets and  simulated Normal and Uniform datasets. Our calculations show both clustering algorithms find higher-occupancy slices than randomly selected slices, and slices of real datasets
have higher occupancy than those from simulated datasets. }
\vspace*{.2cm}
\centering
\begin{tabular}{rrrr}
  \hline
 & random & kmeans & kmed \\ 
  \hline
Decathlon & 8.2 (1.8) & 23.3 (3) & 22.4 (3.7)  \\ 
  Ames & 16.0 (4.3) & 33.6 (8.4) & 25.4 (7.6) \\ 
  Normal & 1.0 (1.0) & 1.8 (0.2) & 1.8 (1.1)\\ 
  Uniform & 1.1 (1.0) & 0.9 (0.1) & 1.3 (1.0)\\ 
   \hline
\end{tabular}
\label{amestab}
\end{table} 
In general, the number of observations visible in sections from real data far exceeds that
from the simulated datasets, as real data tends to be clumpy.
Not surprisingly, paths based on both the clustering methods k-means and k-medoids find sections with many more observations
than simply picking random observations. 

\begin{figure}[htp]
\begin{center}
\includegraphics[scale=.5]{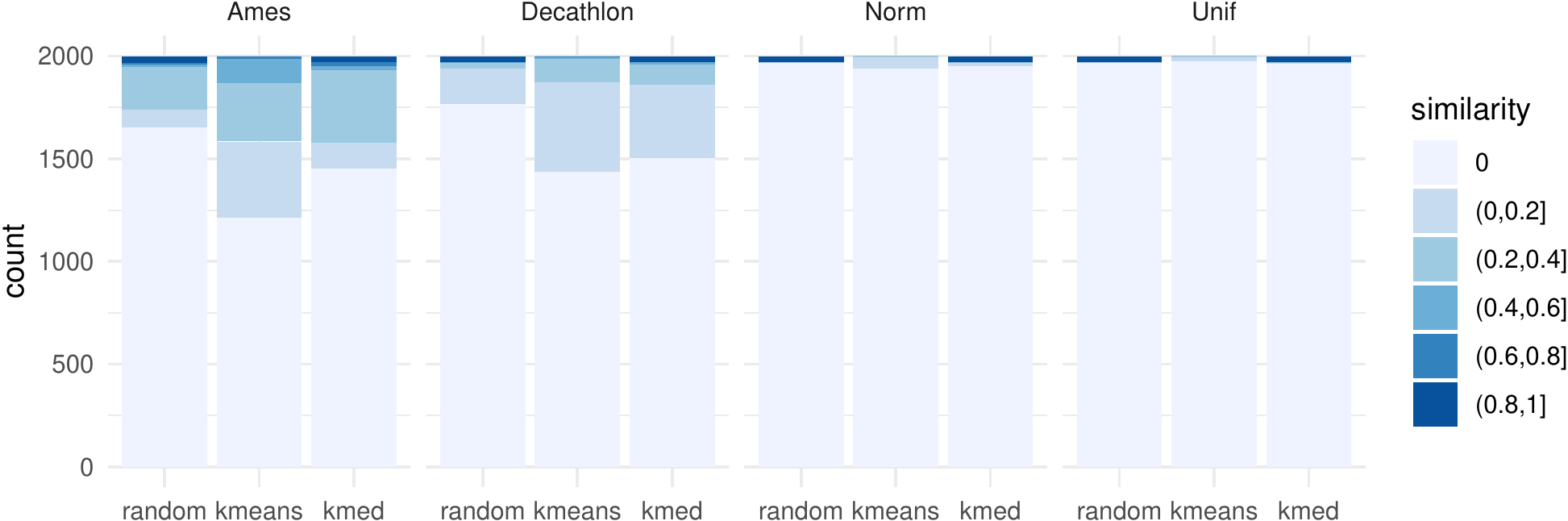}
\end{center}
\caption{Distribution of the maximum similarity per observation across  ($\sigma$=1) maxnorm slices at 30 section points  selected with randomPath, kmeansPath and kmedPath
from Decathlon and Ames datasets and  simulated Normal and Uniform datasets. We see that clustering tours of length 30 visit 25\%  of observations for the real datasets, though not
for the simulated datasets.}
\label{diagnostic}
\end{figure}

We also investigate in Figure \ref{diagnostic}
the distribution of the maximum similarity per observation over the 30 section points for the three path
algorithms and four datasets.
Here, paths based on clustering algorithms from both real datasets visit over 25\% of the observations, again demonstrating that our algorithms 
perform much better on real data than on simulated data.

\subsubsection{Tour construction:  visiting regions exhibiting lack of fit}

Other goals of touring algorithms might be to find regions where the model fits the data poorly,
or where two or more fits give differing results. 
For numeric responses, the tour  lofPath (for lack of fit) finds observations $i$ whose value
of
\begin{align*}
\max_{f \in \rm{fits}} |y_i - \hat{y}^f_{i}|
\end{align*}
is among the $k$ (path length) largest,
where $\hat{y}^f_{i}$ is the prediction for observation $i$ from fit $f$.    
For categorical responses, it finds observations where the predicted class does not match
the observed class.

Another tour called diffitsPath (for difference of fit) finds observations $i$ whose value
of    
\begin{align*}
\max_{f \neq f^{'} \in \rm{fits}} |  \hat{y}^{f^{'}}_{i} -  \hat{y}^f_{i}  |
\end{align*}
is among the $l$ (path length) largest for numeric fits.
For fits to categorical responses,  diffitsPath  currently finds observations where there is the largest
number of distinct predicted categories, or differences in  prediction probabilities.
Other paths could be constructed to identify sections with high amount of fit curvature or
the presence of interaction.

There are a few other simple tours that we have found useful in practice:  tours that visit observations with high and low response values
and tours that move along a selected condition variable, keeping
other condition variables fixed.

\subsubsection{A smoother tour}

For each of the path algorithms, the section points
are ordered using a seriation algorithm to form a
short path through the section points ---dendrogram seriation
\citep{EarleHurley2015} is used here.
If a smoother tour is desired,  the section points $\{ \predictorvals^k_\cond, k=1,2, \ldots, l \}$
may be supplemented with intermediate points formed by  interpolation between $\ \predictorvals^{k}_\cond$
and $ \predictorvals^{k+1}_\cond$.
Interpolation constructs a sequence of evenly
spaced points between each pair of ordered section points. For quantitative predictors, this
means linear interpolation, and for categorical predictors, we simply transition
from one category to the next at the midpoints on the linear scale.

 \section{An interactive implementation}

 The model visualizations on sections and associated touring algorithms described in Section 2 are implemented in our
 highly-interactive R package \pkg{condvis2}.  In the R environment, there are a number of platforms for  building interactive applications.
The most primitive of these is   base R with its function
\code{getGraphicsEvent} which offers control of mouse and keyboard clicks, 
used by our previous package \pkg{condvis} \citep{condvisP, moc}, but the lack of support for other input mechanisms such as menus and sliders
limits the range of interactivity. Tcltk is another option, which is used by
the package \pkg{loon}. We have chosen to use the Shiny platform \cite{shiny}
which is relatively easy to use, provides a browser-based interface and supports web sharing.

First we describe the section plot and condition selector plot panel and the connections between them. These  two displays are combined together with interactive
controls into an arrangement that \cite{ensemble}
refer to as an ensemble layout. 

\subsection{Section plots}
\begin{figure}[htp]
\begin{center}
\begin{tabular}{m{1.2cm}|c|c|}
 & n/f &  p \\
 \hline   
 & (a) & (b)\\
  \raisebox{+20mm}{n/f} 
 &    \raisebox{-8mm}{  \includegraphics[scale=.5,trim=0 0 0 -15]{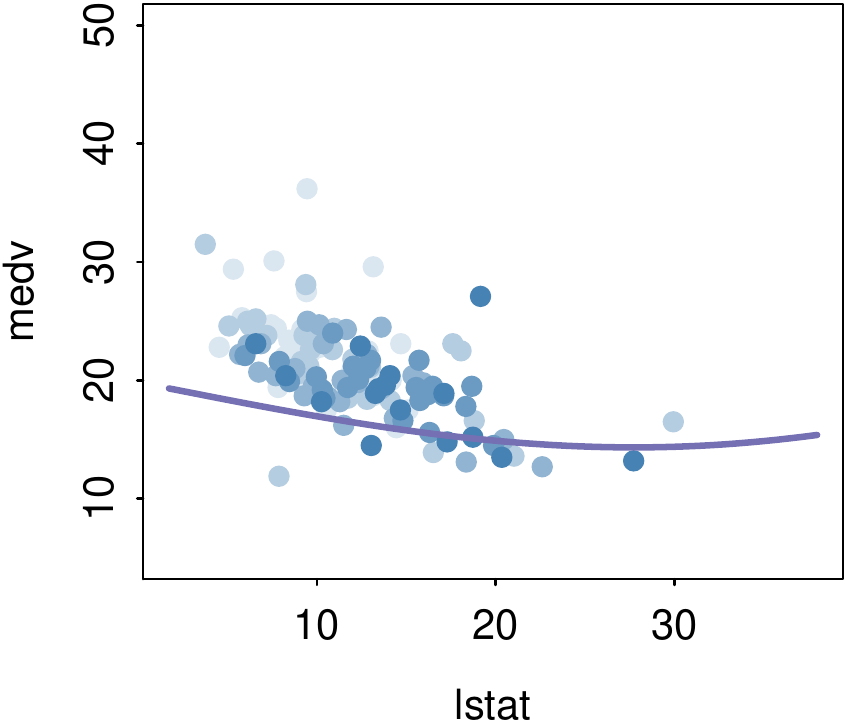} }&  \raisebox{-8mm}{ \includegraphics[scale=.5]{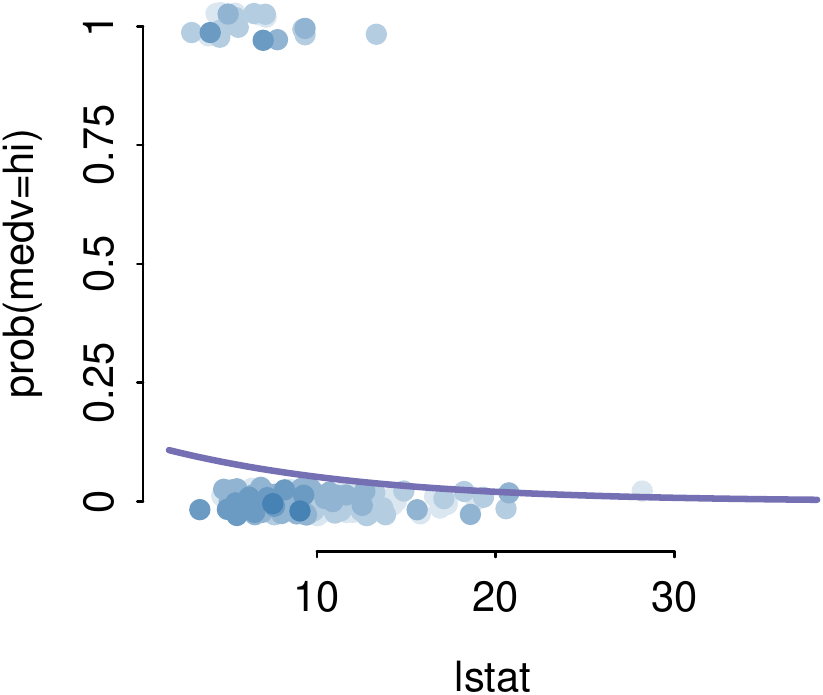}} \\
 \hline 
 & (c) & (d)\\
   \raisebox{+20mm}{(n/f,f)} & \raisebox{-8mm}{\includegraphics[scale=.5,trim=0 0 0 -15]{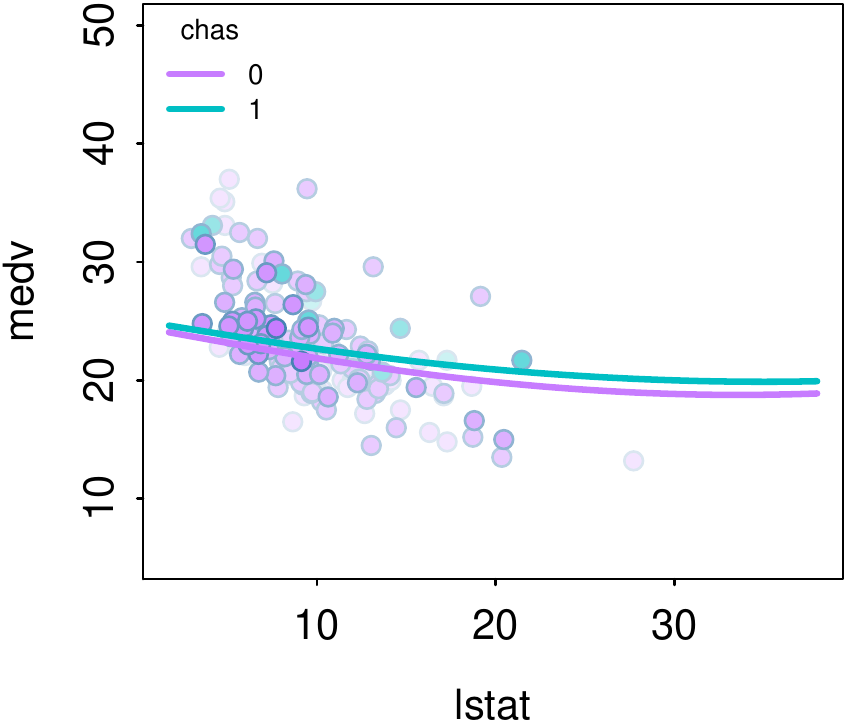}} &  \raisebox{-8mm}{\includegraphics[scale=.5]{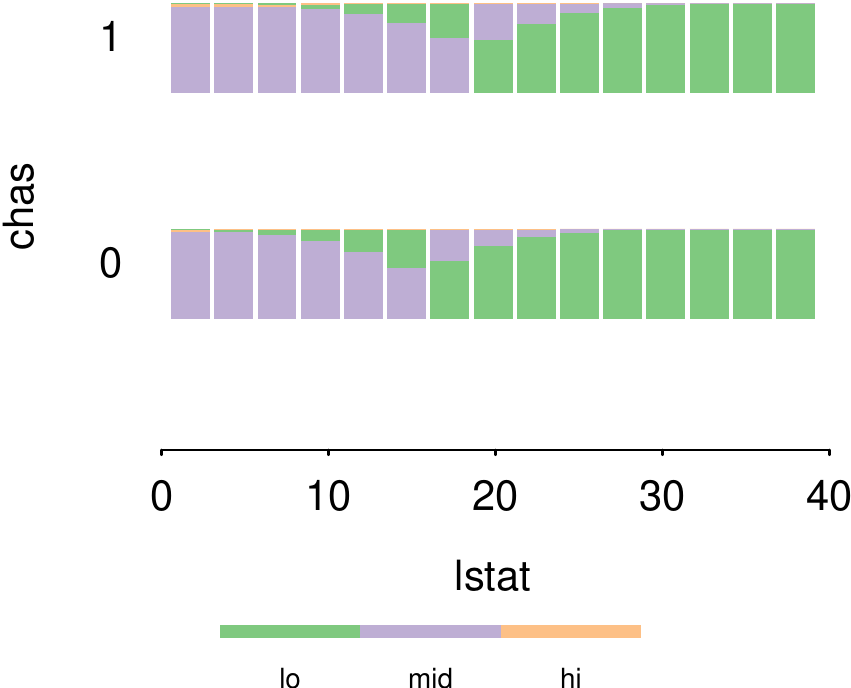}}\\
    \hline
    & (e) & (f)\\
 \raisebox{+20mm}{(n,n)}    & \raisebox{-8mm}{\includegraphics[scale=.5,trim=0 0 0 -15]{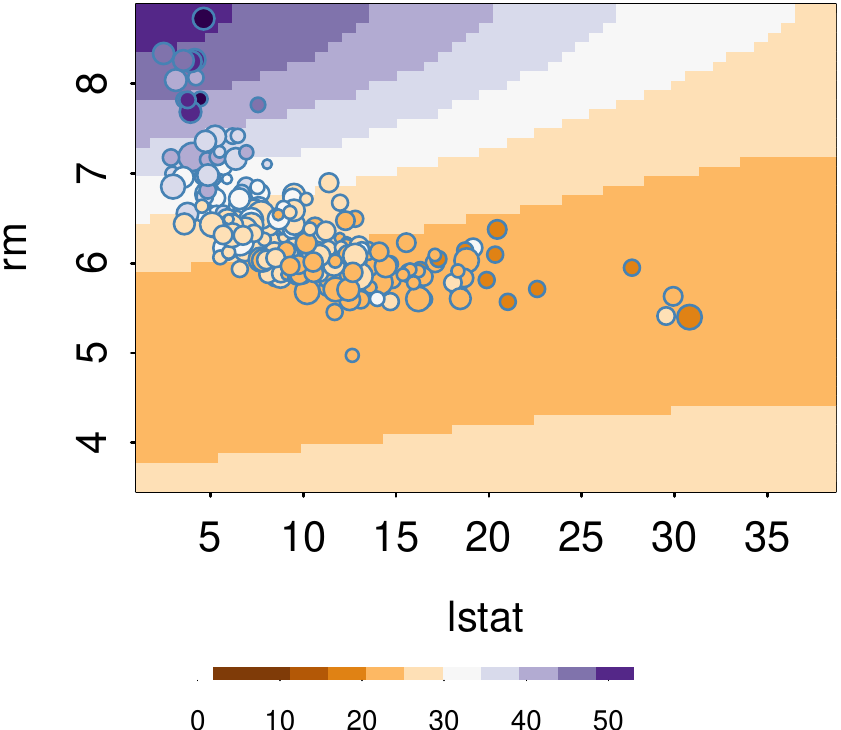}} & \raisebox{-8mm}{\includegraphics[scale=.5]{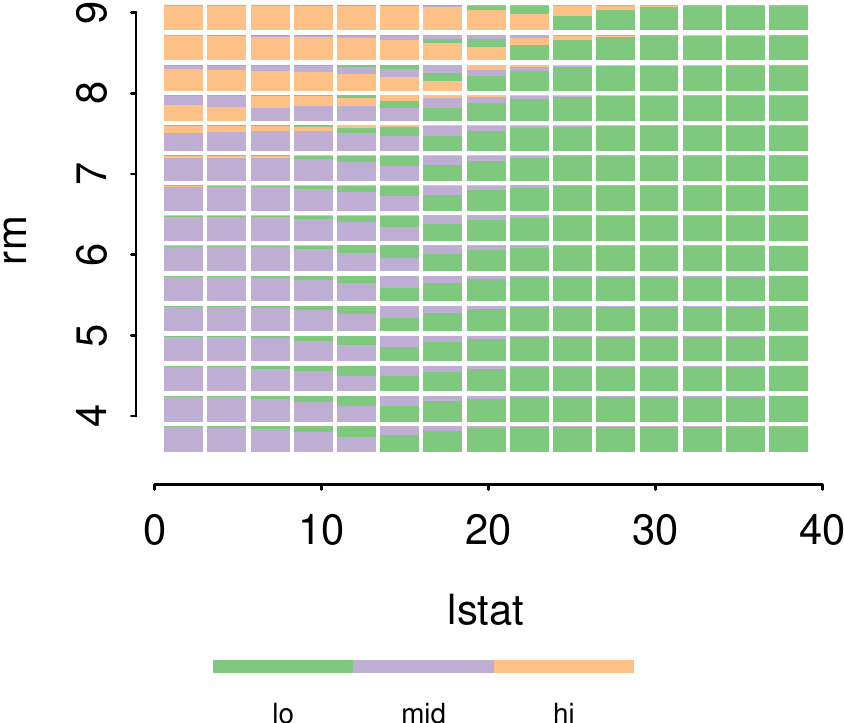}}\\
    \hline 
\end{tabular}
\end{center}
\caption{Types of section plots. Column and row labels represent the prediction and section variable type; n/f: numerical or factor,  p=probability of factor level. 
For n/f types, the factor is treated as numeric.
Plots in the first row have the n/f section variable on the x axis,   y axis has prediction of type  n/f in (a), probability in (b).  
In the second row, there are two section variables, one n/f and the other f, (c) is for a n/f prediction, (d) for multi-class predictions.
In the third row there are two numeric section variables on the axes, (e) is for a n/f prediction shown in color, (f) for multi-class predictions shown as bars.
}
\label{sectionplots}
\end{figure}

As described in Section 2.1, the section plot shows how a fit (or fits) varies with one or two section predictors, for fixed values of the conditioning predictors.
Observations near the fixed values are displayed on the section plot.
A suitable choice of section plot display depends on the prediction (numerical, factor, or probability matrix) and predictor type (numerical or factor).
Figure  \ref{sectionplots}
shows different section plots. For two numeric section variables, we also use perspective displays.
When section predictors are factors these
are converted to numeric, so the displays are similar to those shown in the
rows and columns labelled n/f.
When the prediction is the probability of factor level, the display uses a curve for one of the two levels as in Figure  \ref{sectionplots}(b)
and barplot arrays otherwise, see Figure  \ref{sectionplots}(d) and (f). Here the bars show the predicted class probabilities, for levels of
a categorical section variable, and bins of a numeric section variable. 
For section plot displays such as Figure \ref{sectionplots}(e) where 
fit $\estf$ is shown as an image,  faded points would be hard to see, so instead
we shrink the overlaid observations in proportion to
the similarity score.
We do not add a layer of observations to the barplot arrays in Figure \ref{sectionplots}(d) and (f),
as this would likely overload the plots.

\subsection{Condition selector plots}

The condition selector plots  display  predictors in the conditioning set $C$.  Predictors are plotted singly or in pairs
using scatterplots, histograms, boxplots or barplots as appropriate. They show the distributions of conditioning predictors and also serve as an
input vehicle for new settings of these predictors. 
We use the strategy presented in  \cite{condvis-jss} for ordering 
conditioning predictors to avoid unwitting extrapolation. A pink cross overlaid on the condition selector plots shows the current settings of the section point $\predictorvals_\cond$.
See the panel on the right of Figure \ref{demo}.

Alternatively, predictors may be plotted using
a parallel coordinates display. It is more natural in this setting to restrict conditioning values to observations.
In this case, the  current settings of the section point $ \predictorvals_\cond$ are shown as a highlighted observation.
In principle a scatterplot matrix could  be used, but we do not provide for this option
as it uses too much screen real estate.

\subsection{The condvis2 layout}

\begin{figure}[htp]
\begin{center}
\includegraphics[scale=.54]{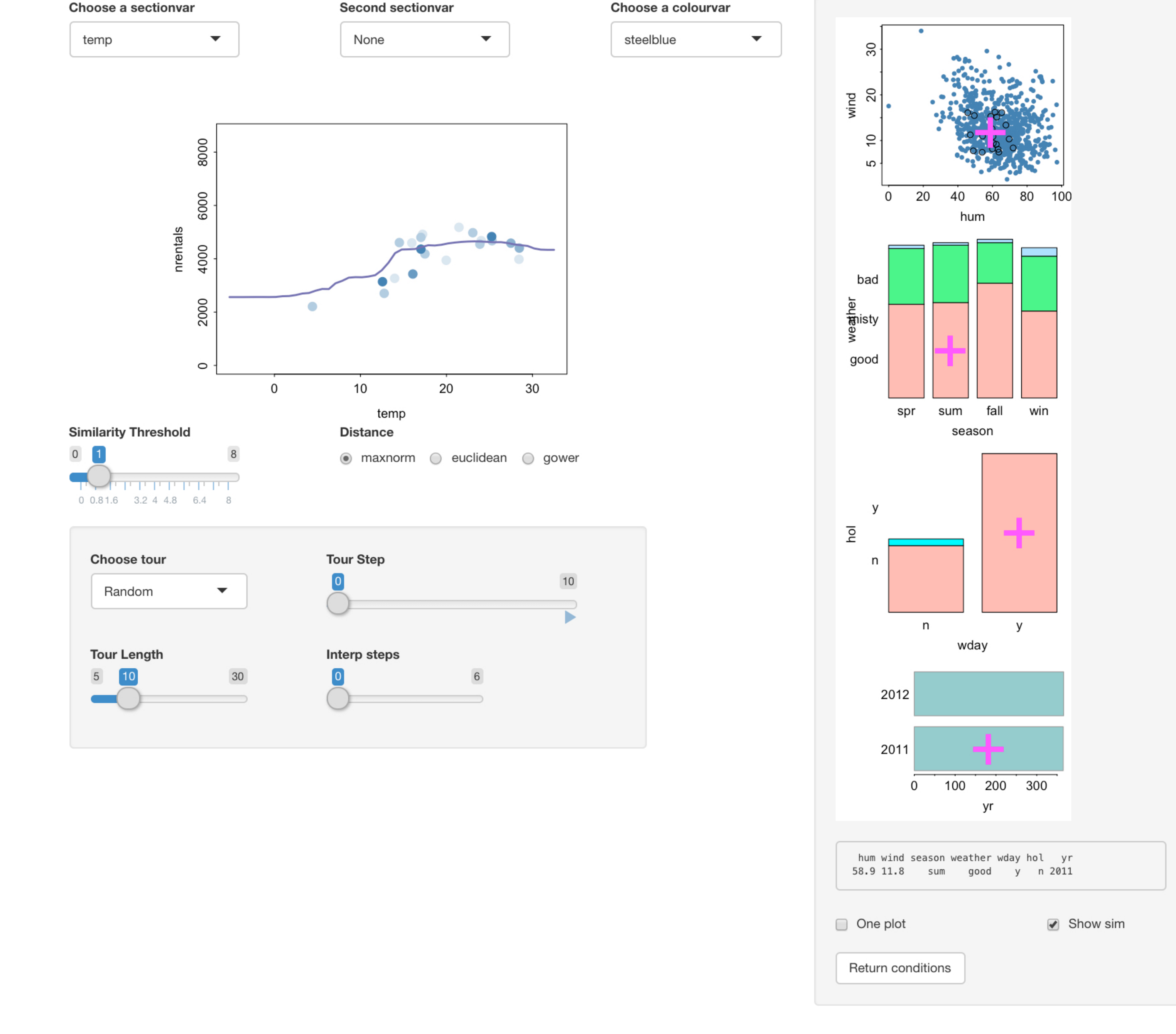}
\end{center}
\caption{Condvis2 screenshot for a random forest fit to the bike rentals data. The nrentals versus temp display is a section plot showing the fit. The panel on the right shows the condition variables, with the current setting marked with a pink cross.
Menus at the top are for selecting section variables, point colors. There is a slider for controlling the similarity threshold, and radio buttons for the distance measure. The bottom left panel is for tour controls.}
\label{demo}
\end{figure}

We introduce a dataset here which we will visit again in Section 4.1.
The bike sharing dataset \citep{bikedata} available from the UCI machine learning repository has a response which is the count of rental bikes (nrentals)
and the goal is to relate this to weather and seasonal information, through features which are
season, hol (holiday or not), wday (working day or not), yr (year 2011 or 2012), weather (good, misty, bad), temp (degrees Celsius), hum (relative humidity in percent)
and wind (speed in km per hour).
The aim is to model the count of rental bikes between years 2011 and 2012 in a bike share system from the corresponding weather and seasonal information.
We build a random forest  \citep{randomforest} fit relating nrentals to other features for all 750 observations.
Setting up an interactive model exploration
requires a call to the function \code{condvis} specifying the data, fit, response, and one or two section variables (here temp).
Other dataset variables become the condition variables.
The resulting  ensemble graphic (see Figure \ref{demo})
has a section plot of  nrentals versus temp  with superimposed random forest fit on the left, the panel on the right has the condition selector plots and the remaining items on the display are
interactive controls.

The pink crosses on the condition selector plots shows the current setting of the conditioning predictors $ \predictorvals_\cond$.
If the initial value of the conditioning predictors is not specified in the call to \code{condvis}, this is set to the medoid of all predictors,
calculated using standardized Euclidean distance, or Gower for predictors of mixed type.
Here $ \predictorvals_\cond$ values are also listed underneath the condition selector plots.  The distance measure used defaults to maxnorm,
so the observations appearing on the section plot all have season=sum, weather=good, wday=y, hol=n, yr=2011, and have wind and hum values
within one (the default value of $\sigma$ in Equation 2) standard deviation of hum=58.0, wind=11.8. The point colors are faded as the maxnorm distance from (hum=58.0, wind=11.8)
increases.  These observations also appear with a black outline on the (hum, wind) condition selector plot.

\subsection{Interaction with condvis2}

The choice of the section point $\predictorvals_\cond$ is under interactive control. 
The most direct way of selecting $\predictorvals_\cond$ is by interacting with the 
condition selector plots,
For example, clicking on the (hum, wind) plot in Figure \ref{demo} at location 
 (hum=90, wind=10) moves the coordinates of $\predictorvals_\cond$ for these two variables to the new location, 
 while the values for other predictors in $C$ are left unchanged. Immediately the section plot  of nrentals versus temp shows the random forest
 fit at the newly specified location, but now there is only one observation barely visible in the section plot, telling us that the current combination of the conditioning predictors
 is in a near-empty slice.
Double-clicking on the (hum, wind) plot sets the section point to the closest observation on this plot. 
If there is more than one such observation, then the section point becomes  the medoid of these closest observations.
It is also possible to click on an observation in the section plot, and this has the effect of moving the section point 
 $\predictorvals_\cond$ to the coordinates of the selected observation for the conditioning predictors.

The light grey panel on the lower left has the tour options (described in Section 2.2)
which
 offer another way of navigating slices of predictor space.
 The  ``Choose tour'' menu offers  a choice of tour algorithm,  and ``Tour length''
controls the length of the computed path.
The ``Tour Step'' slider controls the position along the current path; by clicking the arrow on the right the tour
progresses automatically through the tour section points.
An interpolation option is available for smoothly changing paths.

 Clicking on the similarity threshold slider increases or decreases the value of $\sigma$, including more or less observations in the nrentals versus temp plot.
 The distance used for calculating similarities may be changed from maxnorm to Euclidean or Gower (see Equations 3 and 4) via the radio buttons. When the threshold slider
 is moved to the right-most position, all observations are included in the section plot display.

 One or two  section variables may be selected from the ``Choose a sectionvar'' and  ``Second sectionvar'' 
 menus. If the second section variable is hum, say, this variable is removed from the condition selector plots.
 With two numeric section variables, the section plot appears as an image  as in Figure \ref{sectionplots}(e). Another checkbox  
``Show 3d surface'' appears, and clicking this shows how the fit relates to (temp, hum) as a rotatable 3d plot.
 Furthermore, 
 a variable used to color observations may be chosen  from the ``Choose a colorvar'' menu. 

 Clicking the ``One plot'' checkbox on the lower right changes the condition selector plots to a single parallel coordinate plot.
 Deselecting the ``Show sim'' box causes the black outline on the observations in the current slice to be removed, which is a useful
 option if the dataset is large and display speed is an issue.
 Clicking on the ``Return conditions'' button causes the app to exit, returning all section points visited as a data frame.

\subsection{Which fits?}
Visualizations in \pkg{condvis2} are constructed in a model-agnostic way.
In principle all  that is required is that a fit produces predictions. Readers familiar with R
will know that algorithms from random forest to logistic regression to support vector machines
all have some form of \code{predict} method, but they have different arguments and interfaces.

We have solved this  by writing a predict wrapper called \code{CVpredict} (for condvis predict)
that operates in a consistent way for a wide range of fits.
We provide over  30 \code{CVpredict} methods, for fits ranging from neural nets, to trees to bart machine.
And, it should be relatively straightforward for others to write their own \code{CVpredict}  method, using the template 
we provide.

Others have tackled the problem of providing a standard interface to the model fitting
and prediction tasks. The \pkg{parsnip} package  \citep{parsnip}  part of the so-called tidyverse world
streamlines the process and currently includes drivers for about 40 supervised learners
including those offered by spark and stan.
The packages \pkg{caret}  \citep{caret},  \pkg{mlr} \citep{mlr}, and its  most recent incarnation \pkg{mlr3} \citep{mlr3}, interface with hundreds of learners and also support parameter tuning.
As part of  \pkg{condvis2}, we have written \code{CVpredict} methods for the model fit classes 
from \pkg{parsnip}, \pkg{mlr}, \pkg{mlr3} and \pkg{caret}.
 Therefore our  visualizations
are accessible from fits produced by most of R's machine learning algorithms.

\subsection{Dataset size }

Visualization of large datasets is challenging, particularly so in interactive settings where a user expects
 near-instant response. We have used our application in settings with $n= 100,000$ and $p=30$ and the computational
burden is manageable.

For section displays, the number of points displayed is controlled by the similarity threshold $\sigma$ and
is usually far below the dataset size $n$.
For reasons of efficiency, condition selector displays by default show at most 1,000 observations, randomly selected in
the case where $n>1000$. 
Calculation of the medoid for the initial section point and the kmedPath  requires calculation of a distance matrix
which has complexity $O(n^2 p)$. For interactive use speed is more important than accuracy so we base these calculations on a maximum of 4,000
rows by default.

The conditioning displays show $\lceil p/2 \rceil$ panels of one or two predictors or one parallel coordinate display.
Up to  $p=30$ will fit on screen space using  the  parallel coordinate display, perhaps 10-15 otherwise.
Of course many data sets have much larger feature sets. In this situation, we recommend selecting
a subset of features which are \emph{important} for prediction, to be used as the section and conditioning
predictors $\sect$ and $\cond$. The remaining set of predictors, say $F$, are hidden from view in the
condition selector plots and are fixed at some initial
value which does not change throughout the slice exploration.

Note that though the predictors $F$ are ignored in the calculation of
distances in Equations  \ref{minkowski} and \ref{gower}  and thus in the similarity scores of
Equation  \ref{eq:simweight}, the initial values of these predictors $ \predictors_F = \predictorvals_F$ are used throughout  in
constructing predictions; thus the section plot shows
$\estf(\predictors_\sect = {\predictors_\sect}^g, \predictors_\cond = \predictorvals_\cond, \predictors_F = \predictorvals_F)$.
 If the set of important predictors is not carefully selected,
the fit displayed will not be representative of the fit for all observations visible in the section plot.

In the situation where some predictors  designated as unimportant are
relegated to $F$ thus not appearing in the condvis display, the settings for predictors in $F$ remain at their initial values throughout all tours.
This means that section points for the tours based on selected observations 
(randomPath, kmedPath, lofPath and  diffitsPath) will not in fact correspond 
exactly to dataset observations. 
An alternative strategy would be to let the settings for the predictors in $F$  vary, but then
there is a danger of being ``lost in space''.

\section{Applications }

In our first example, we compare a linear fit with a random forest for a regression problem. Interactive exploration leads us to discard the linear fit
as not capturing feature effects in the data, but patterns in the random forest fit suggests a particular generalized additive model
that overall fits the data well.

Our second example concerns a classification problem where we compare  random forest and tree fits. We learn that
both fits have generally similar classification surfaces. In some boundary regions the random forest overfits the training data
avoiding the mis-classifications which occur for the tree fit.

Finally, we review briefly how interactive slice visualization techniques can be used in unsupervised learning problems, 
namely to explore density functions and estimates, and clustering results.
Furthermore, we demonstrate that interactive slice visualization is insightful even in situations where there is no fit curve or surface to be plotted.

\subsection{Regression: Bike sharing  data} 
\begin{figure}[htp]
\begin{center}
\includegraphics[scale=.5,trim={0 0 10 10 },clip]{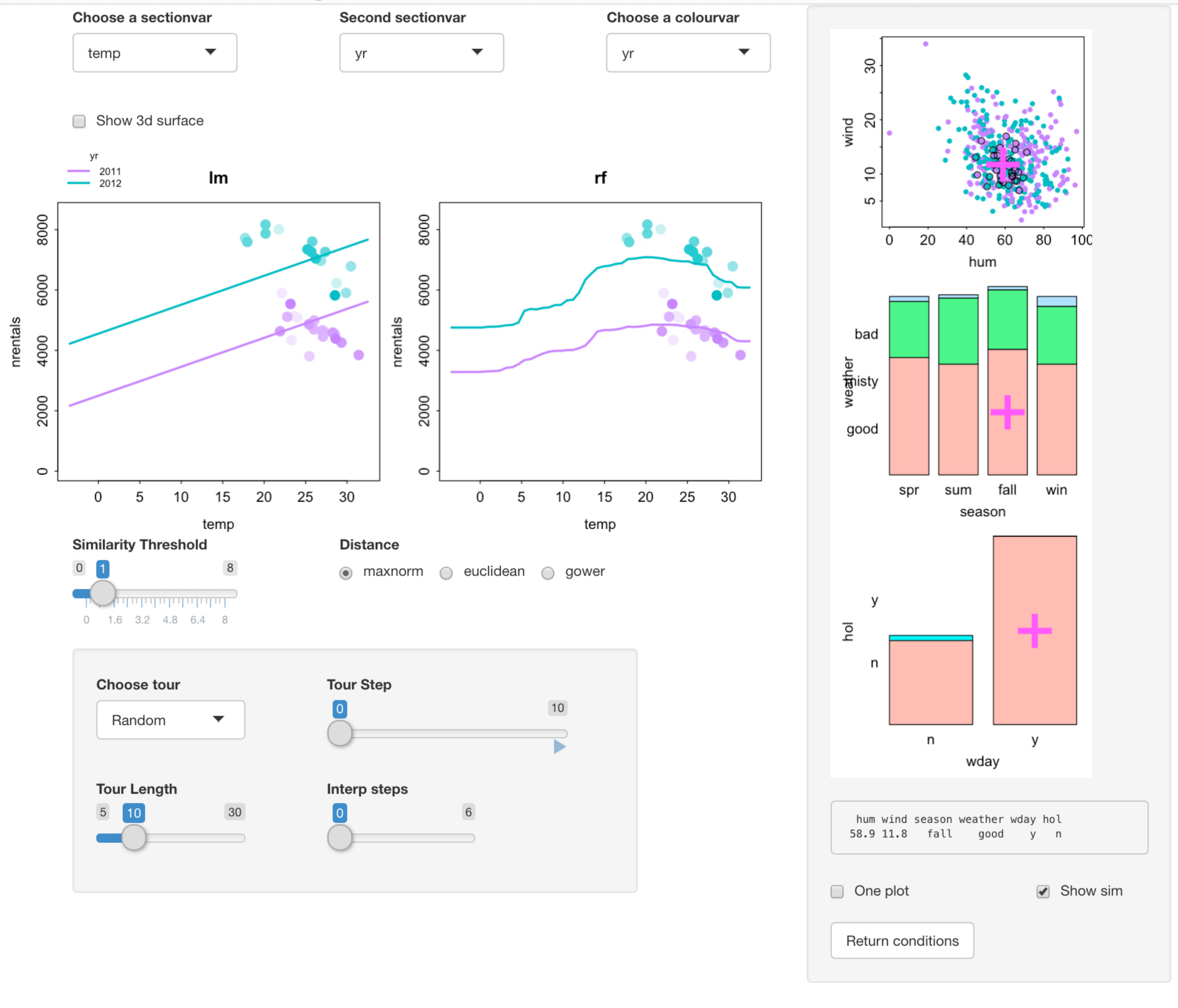}
\end{center}
\caption{The random forest and linear model fit for the bike rentals training data. The section variables are temp and year. The linear model fits poorly. The random forest
has a decreasing trend for both years for temperatures above 15C, which is supported by nearby observations.}
\label{bikerevslm}
\end{figure}

Here we  investigate predictor effects and goodness of fit for models fit to the  bike sharing dataset, introduced in Section 3.3.
To start with, we divide the data into training and testing sets using a 60/40 split.
For the training data, we fit a linear model with no interaction terms, and  a random forest which halves the RMSE by comparison with the linear fit.
Comparing the two fits we see that the more flexible fit is much better supported by the data, see for example Figure \ref{bikerevslm}.
In the fall, bike rentals are affected negatively by temperature according to the observed data. The linear fit does not pick up this trend, and
even the random forest seems to underestimate the effect of temperature.
Year is an important predictor: people used the bikes more in 2012 than in 2011.   At the current setting of the condition variables, there is no data below a temperature of 15C,
so we would not trust the predictions in this region.

\begin{figure}[htp]
\begin{tabular}{ccc}
\includegraphics[scale=.35,trim={50 200 100 220},clip]{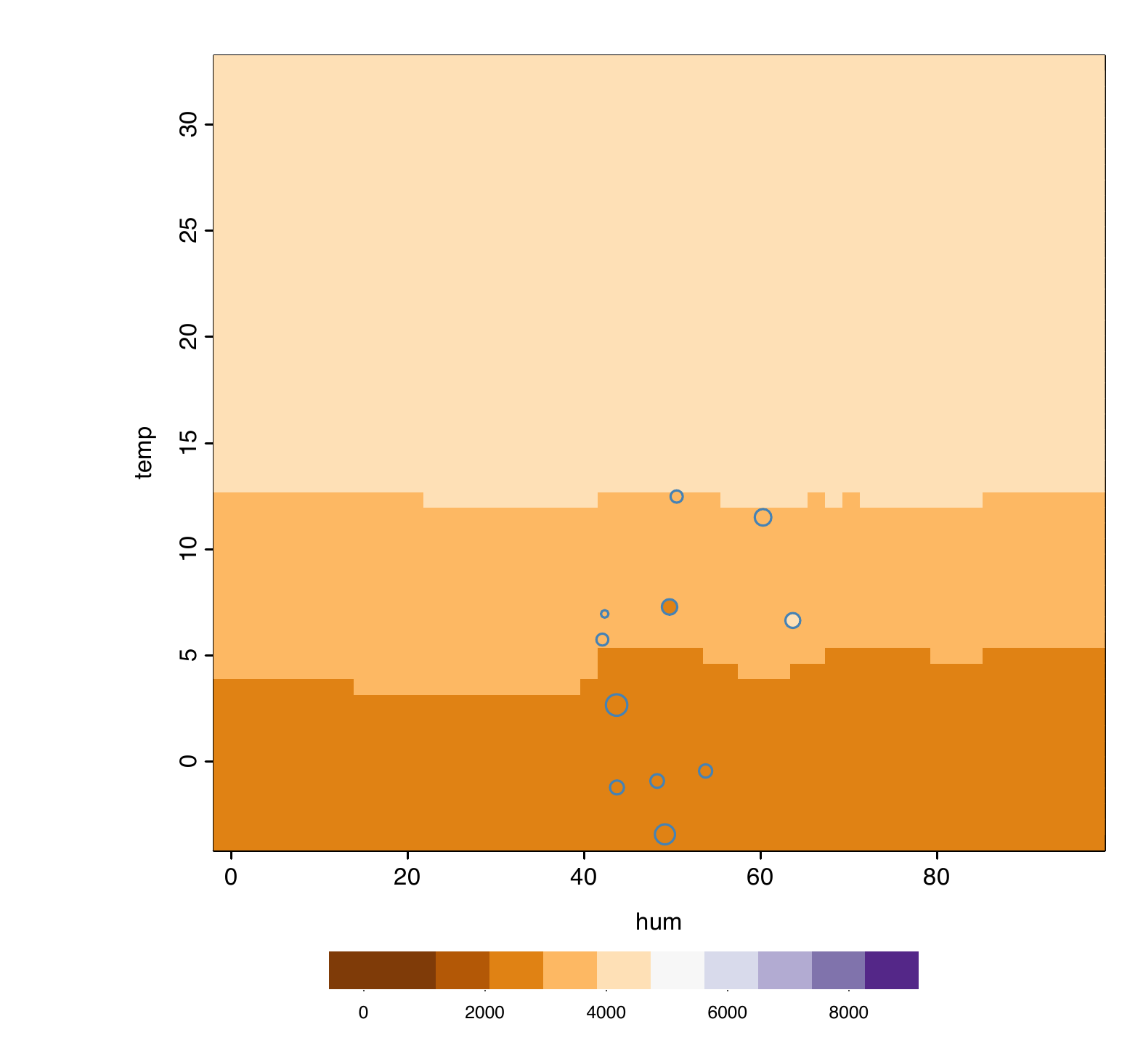} & \includegraphics[scale=.35,trim={134 200 100 220},clip]{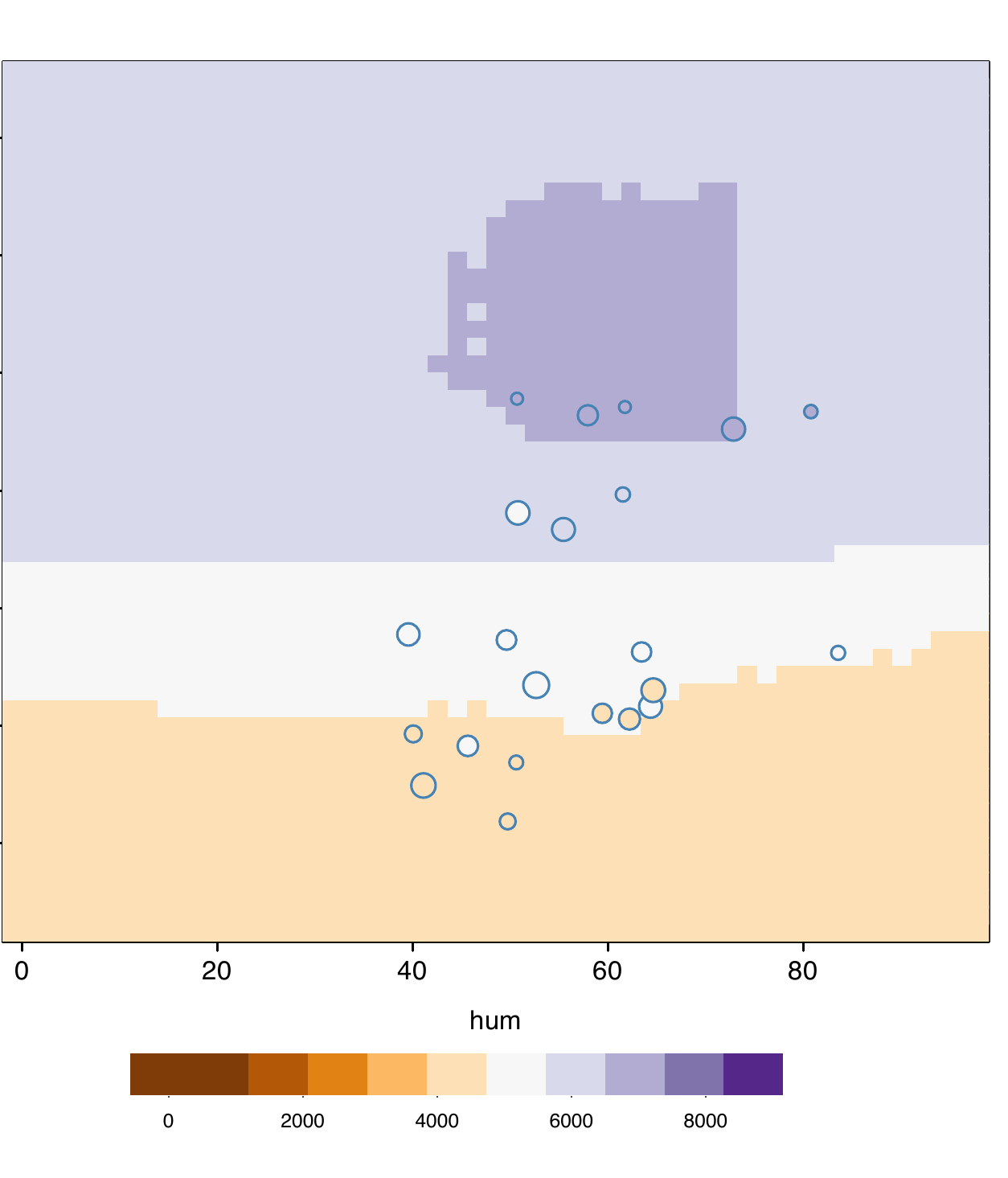} & \includegraphics[scale=.35,trim={134 200 100 220},clip]{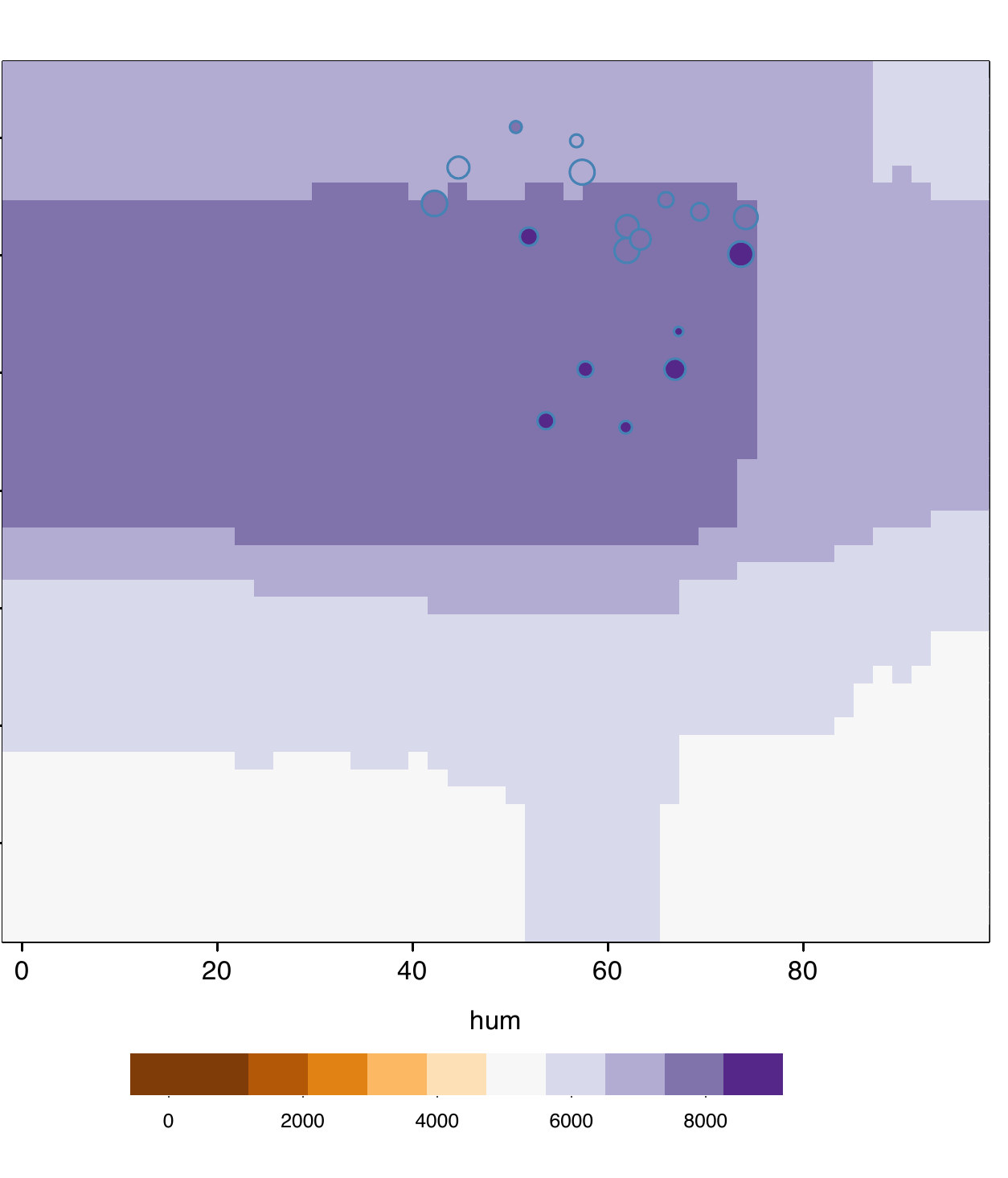}\\
 (a) Spring 2011 & (b) Spring 2012 & (c) Fall 2012
  \end{tabular}
\caption{Section plots with  predictors temperature and humidity of  random forest fit to bike training data. Image  color shows the predicted number of rentals. 
Conditioning variables other than year and season are set to good weather/weekend/no holiday.
Comparing the three plots, we see that the joint effect of humidity and temperature  changes through time; that is,  a three-way interaction.}
\label{bikerf1}
\end{figure}

Focusing on the random forest only, we explore the combined effect on nrentals of the two predictors temperature and humidity) (Figure \ref{bikerf1}). 
The three plots have different settings of the time condition variables selected interactively, other conditioning variables were set to good weather, weekend and no holiday.
In spring 2011, temperature is the main driver of bike rentals, humidity has negligible impact. In spring 2012 the number of bike rentals is higher than the previous year, especially at higher temperatures.
In fall 2012, bike rentals are higher than in spring, and high humidity reduces bike rentals.
With further interactive exploration, we see that this three-way interaction effect is consistent at other levels of weather, weekend and holiday.

\begin{figure}[htp]
\begin{center}
\includegraphics[scale=.4]{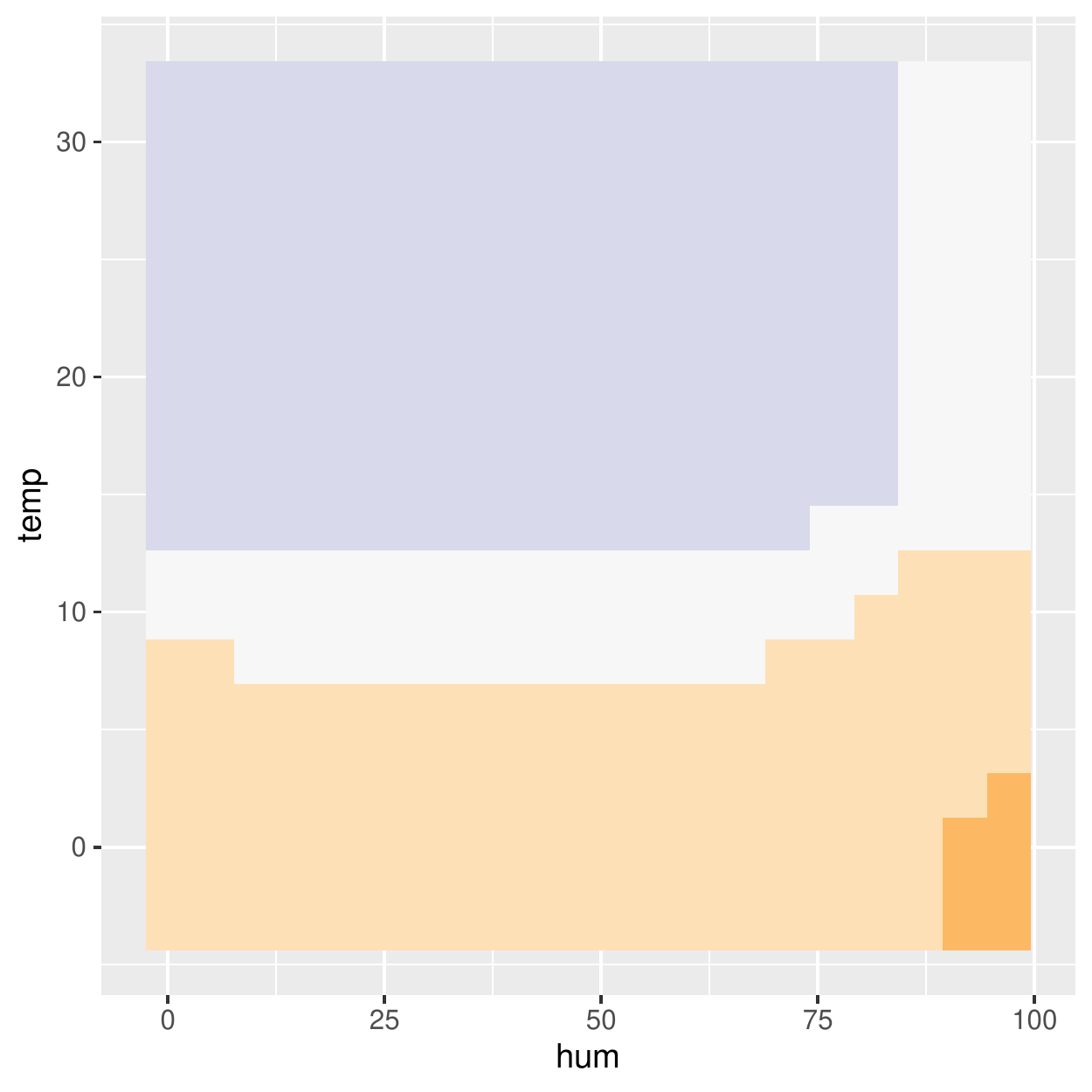}\\
\includegraphics[scale=.4,trim={0 0 0 240 },clip]{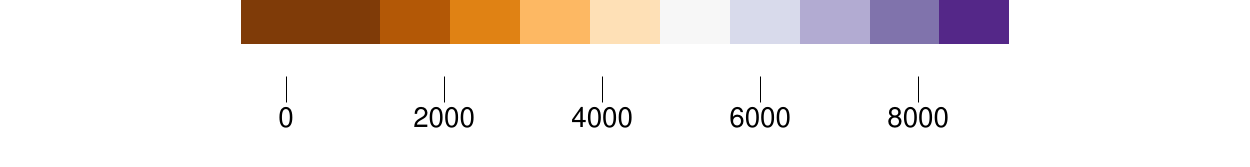}
\end{center}
\caption{Partial dependence plot for random forest fit to bike training data, showing effect of temperature and humidity on the predicted number of rentals. The plot shows an interaction effect: prediction is higher for temperature above 12C,
but drops off for humidity above 80.}
\label{bikePDP}
\end{figure}

In the absence of an interactive exploratory tool such as ours, one might summarize the joint effect of temperature and humidity through a partial dependence plot (Figure \ref{bikePDP}).
The  plot  combines the main effect of the featuress and their interaction effect, and shows that people cycle more when temperature is above 12C, and this effect depends on humidity. The partial dependence plot is a summary of
plots such as those in Figure \ref{bikerf1}, averaging over all observations in the training data for the conditioning variables, and so it cannot
uncover a three-way interaction. A further issue is that  the partial dependence curve or surface is averaging over fits which are extrapolations, 
leading to conclusions which may not be reliable.

Based on the information we have gleaned from our interactive exploration, an alternative parametric fit to the random forest is suggested.  We build a generalized additive model (gam),
with a smooth joint term for temperature, humidity, an interaction between temperature and season, a smooth term for wind, and a linear term
for the remaining predictors. A gam fit is parametric and will be easier to understand and explain than a random forest, and has the additional advantage of providing confidence intervals, which may be added to the condvis2 display.
Though the training RMSE for the random forest is considerably lower than that for the gam,
on the test data the gam is a clear winner, see Table \ref{rmse}.
\begin{table}[htp]
\centering
\caption{Training and test RMSE for the random forest and gam fits to the bike data. The gam has better test set performance than the random forest.} 
\begin{tabular}{rrr}
  \hline
 & train & test \\ 
  \hline
RF & 438.1 & 748.1  \\ 
  GAM & 573.1 & 670.2 \\ 
   \hline
\end{tabular}

\label{rmse}
\end{table}

\begin{figure}[htp]
\includegraphics[trim={0 0 0 0},clip,scale=.363 ]{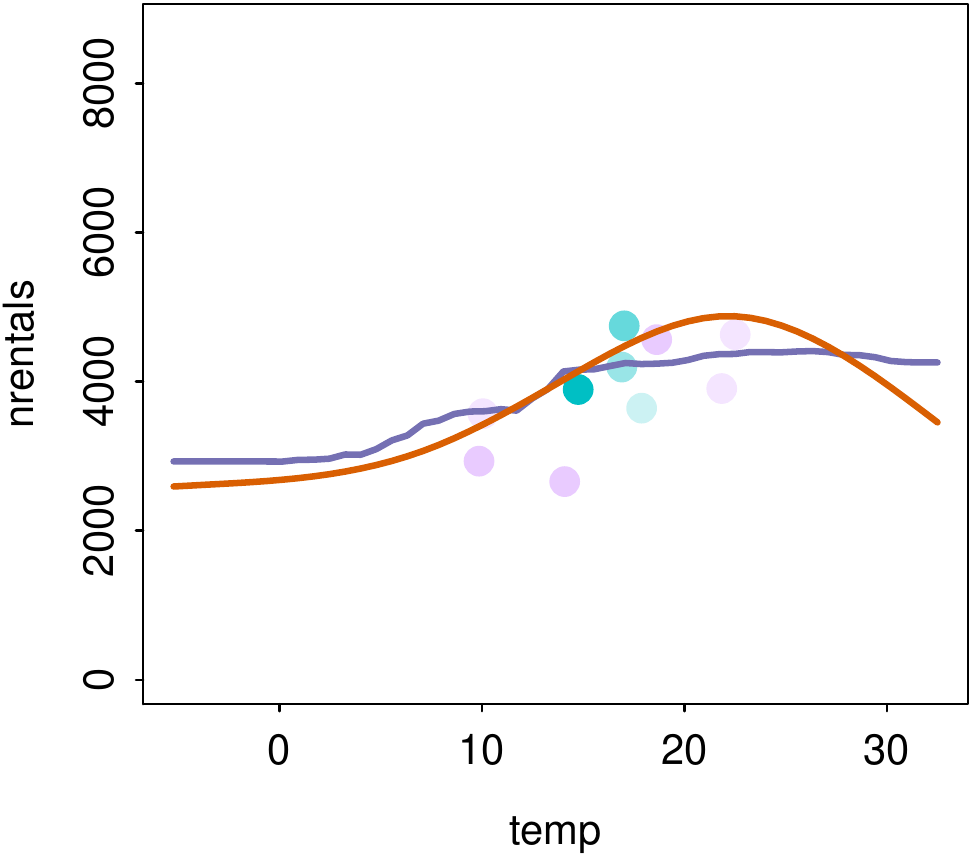} 
\includegraphics[scale=.363,trim={41 0 0 0},clip ]{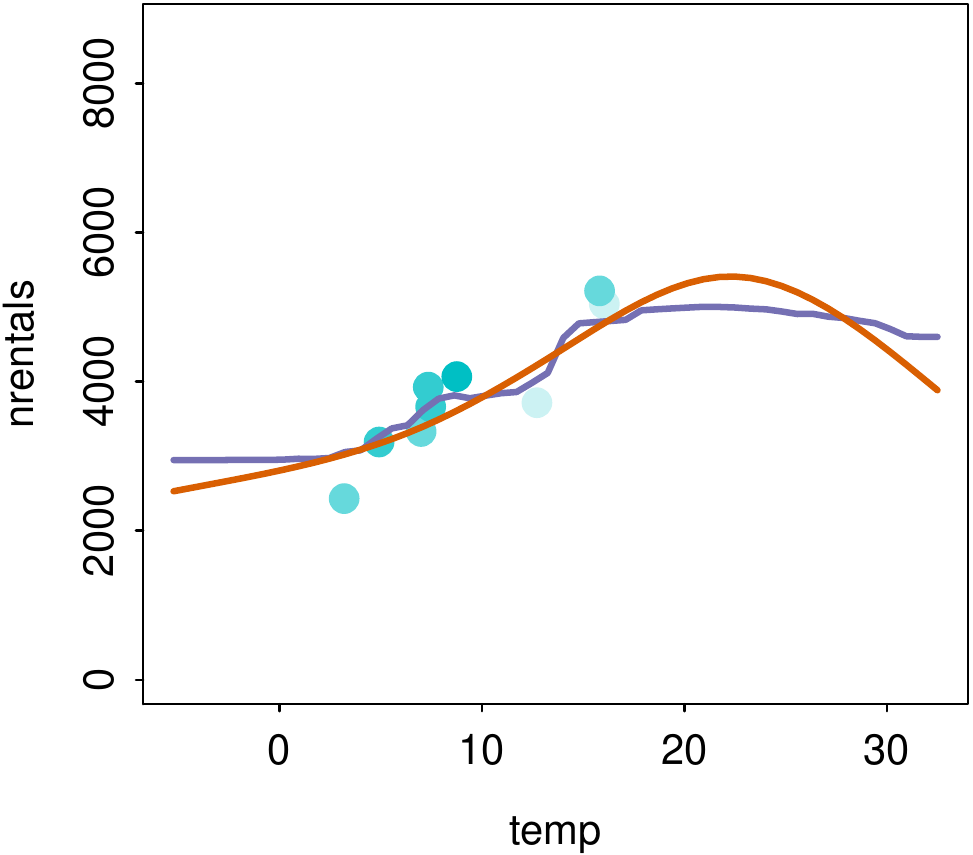} 
\includegraphics[scale=.363,trim={41 0 0 0},clip ]{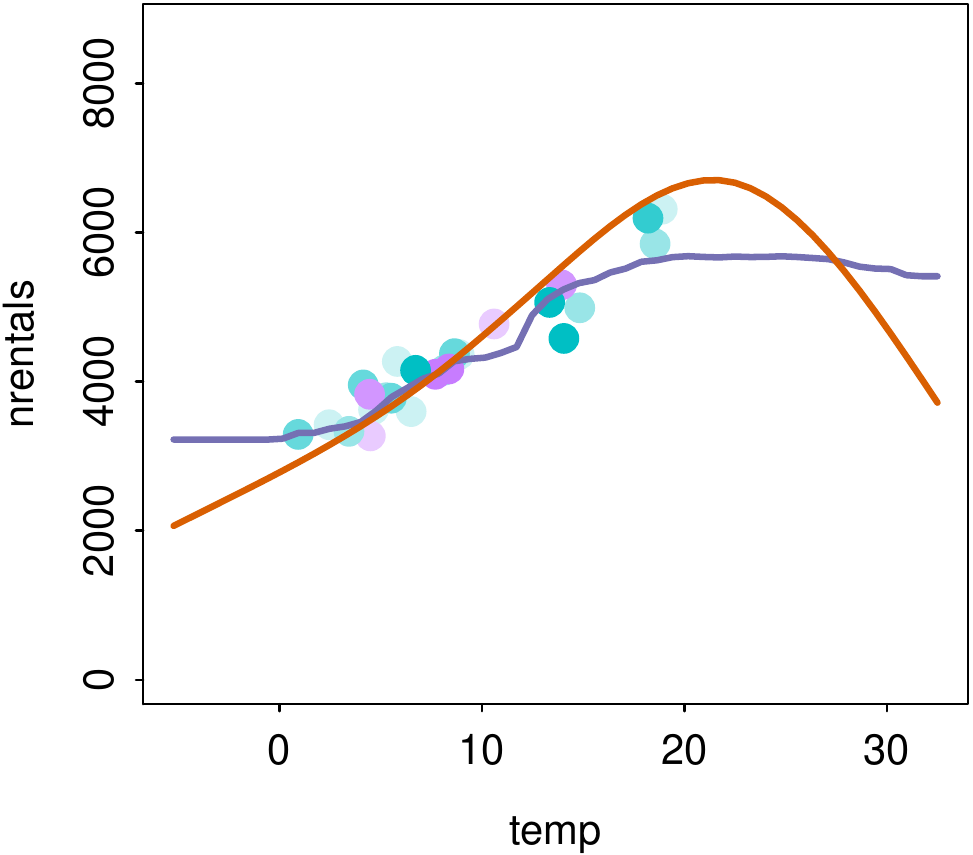} 
\includegraphics[scale=.363,trim={41 0 0 0},clip ]{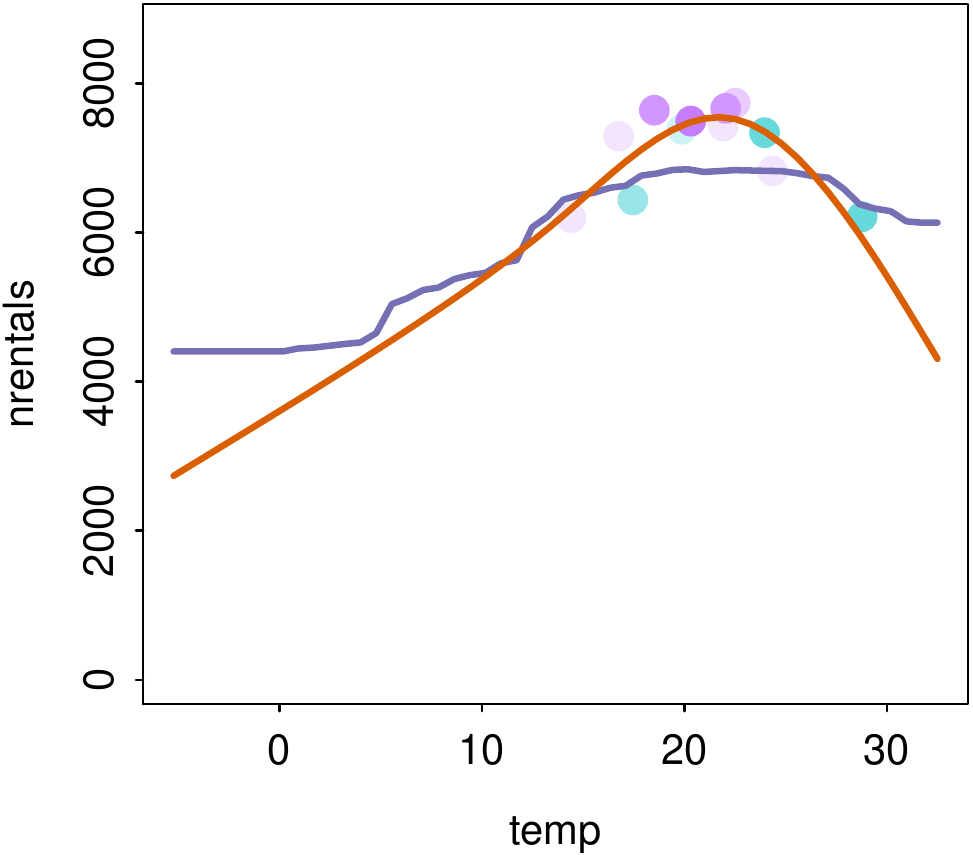} 
\includegraphics[scale=.363,trim={41 0 0 0},clip ]{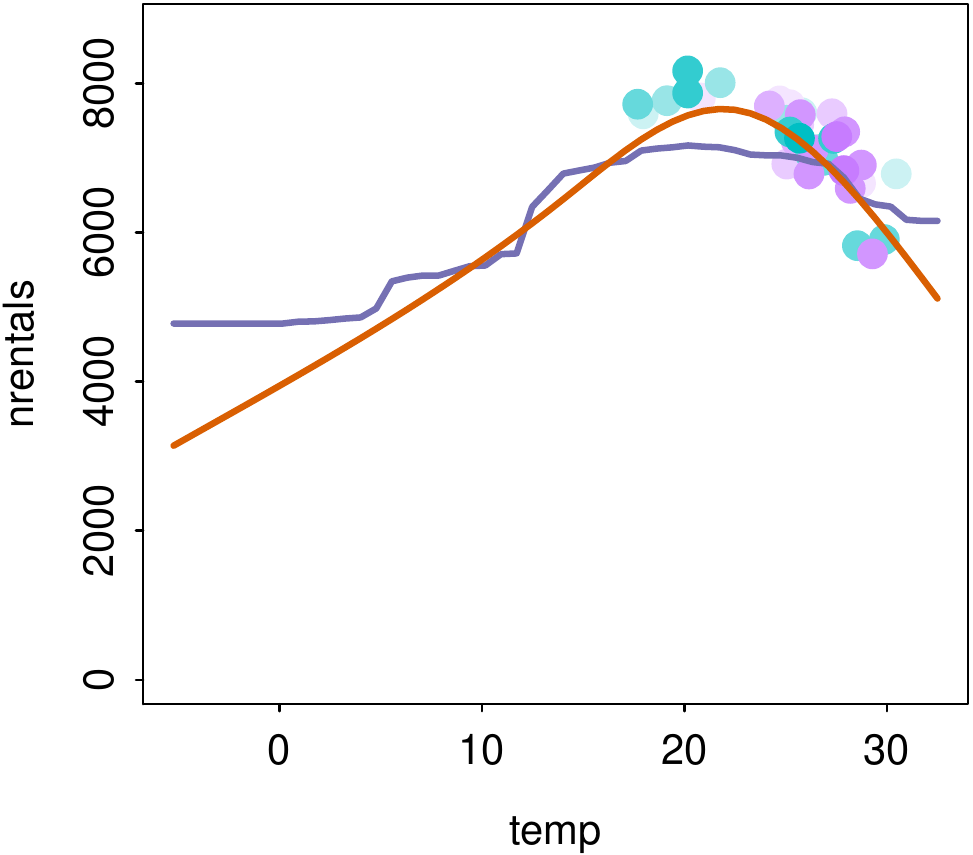} 
\includegraphics[scale=.363,trim={0 0 0 0},clip ]{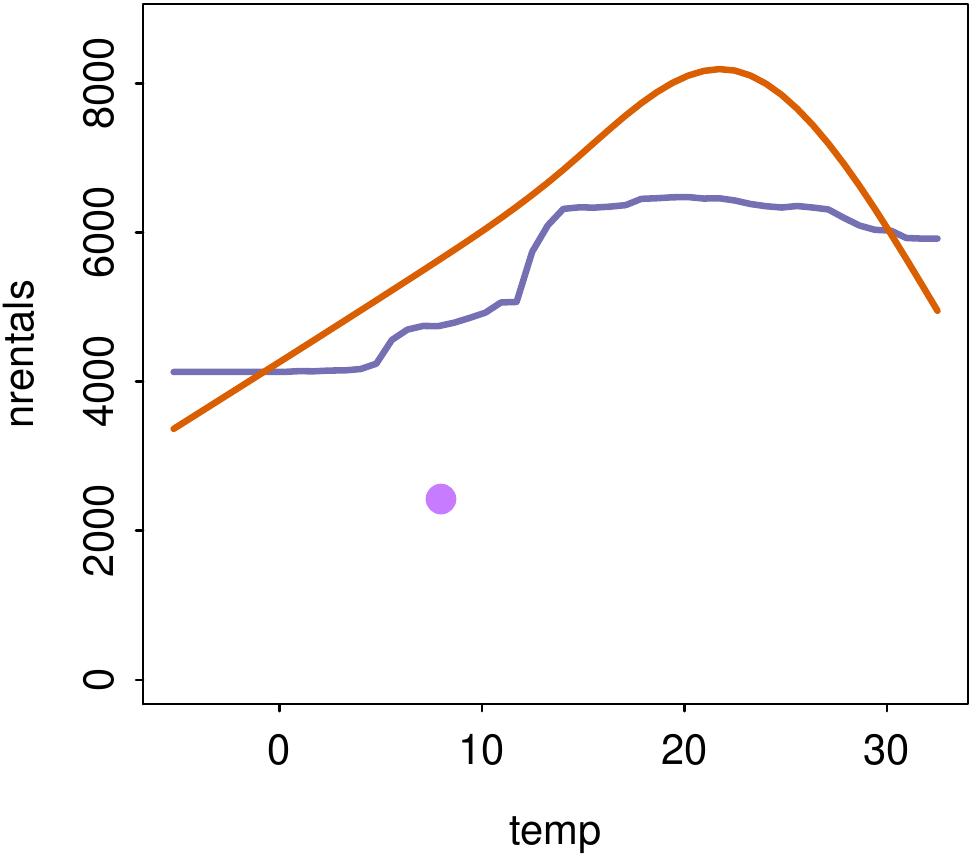} 
\includegraphics[scale=.363,trim={41 0 0 0},clip ]{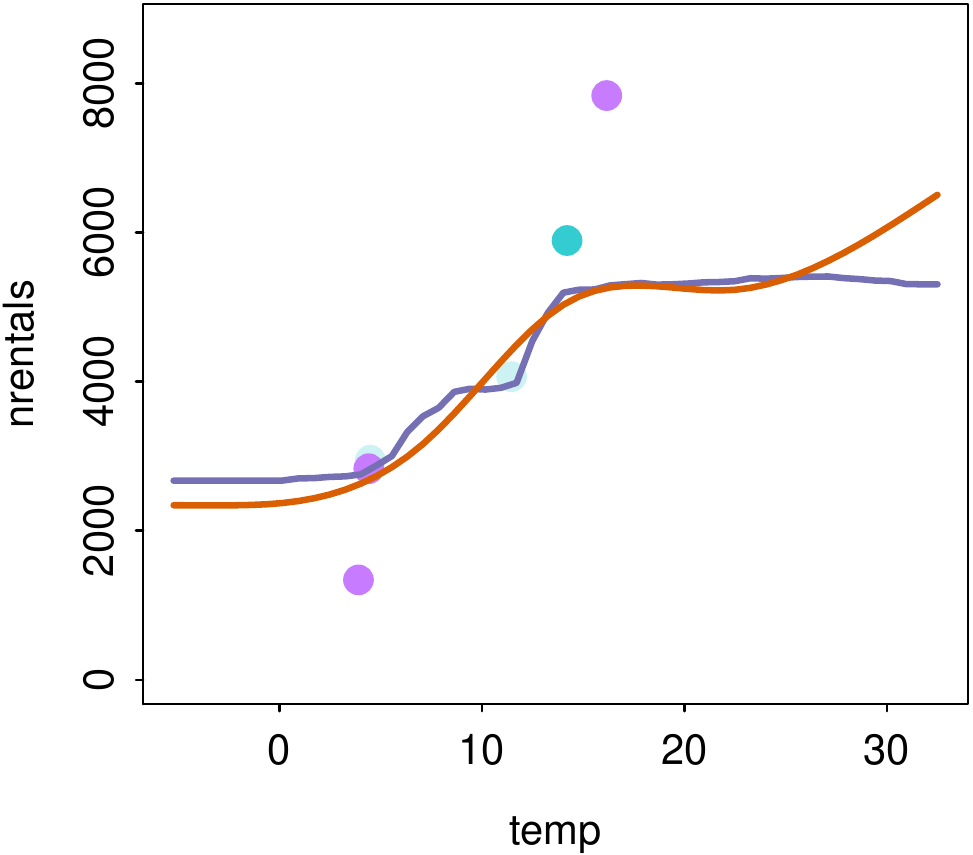} 
\includegraphics[scale=.363,trim={41 0 0 0},clip ]{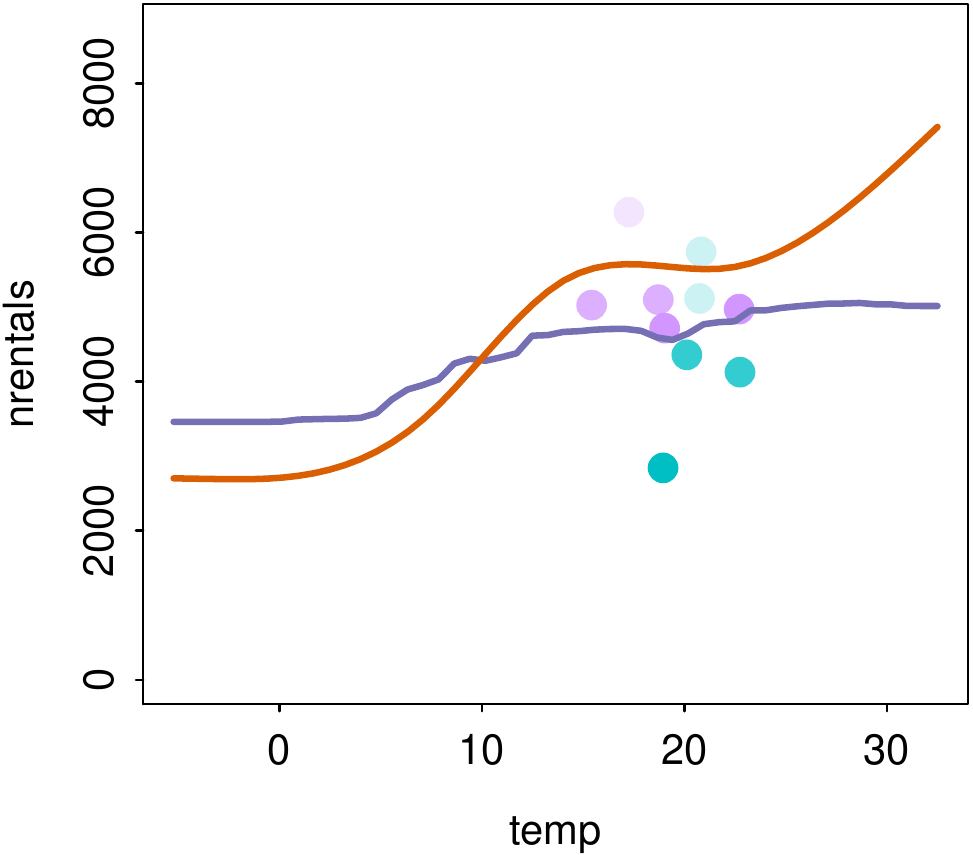} 
\includegraphics[scale=.363,trim={41 0 0 0},clip ]{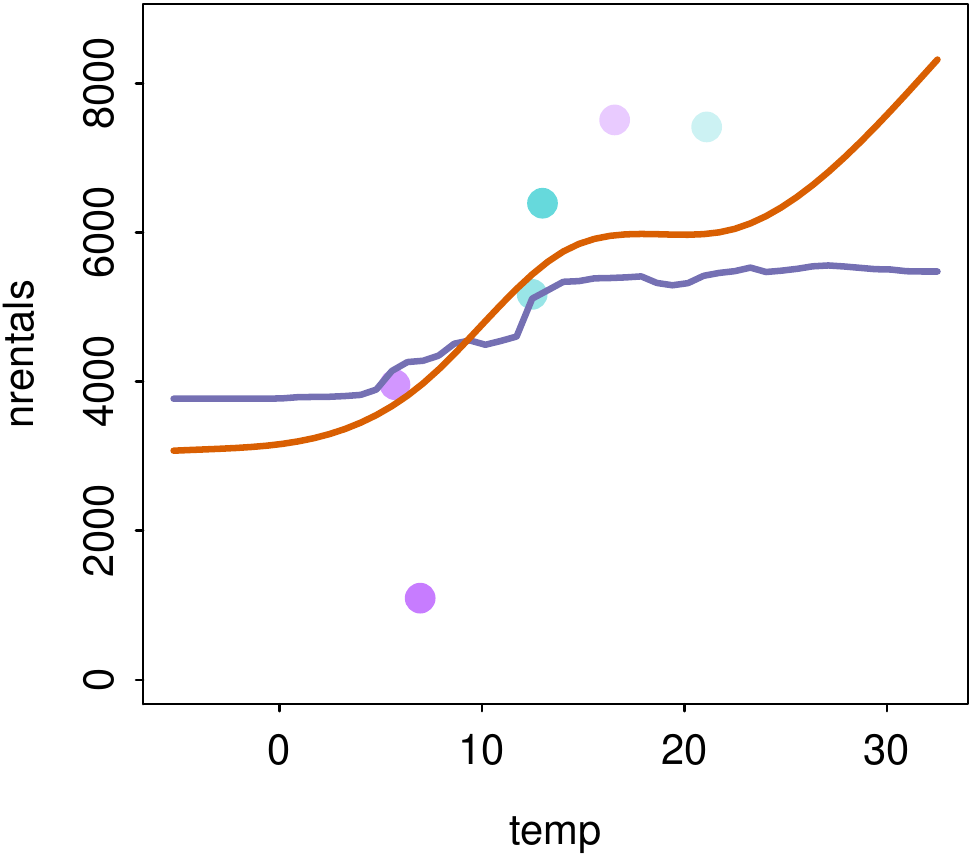} 
\includegraphics[scale=.363,trim={41 0 0 0},clip ]{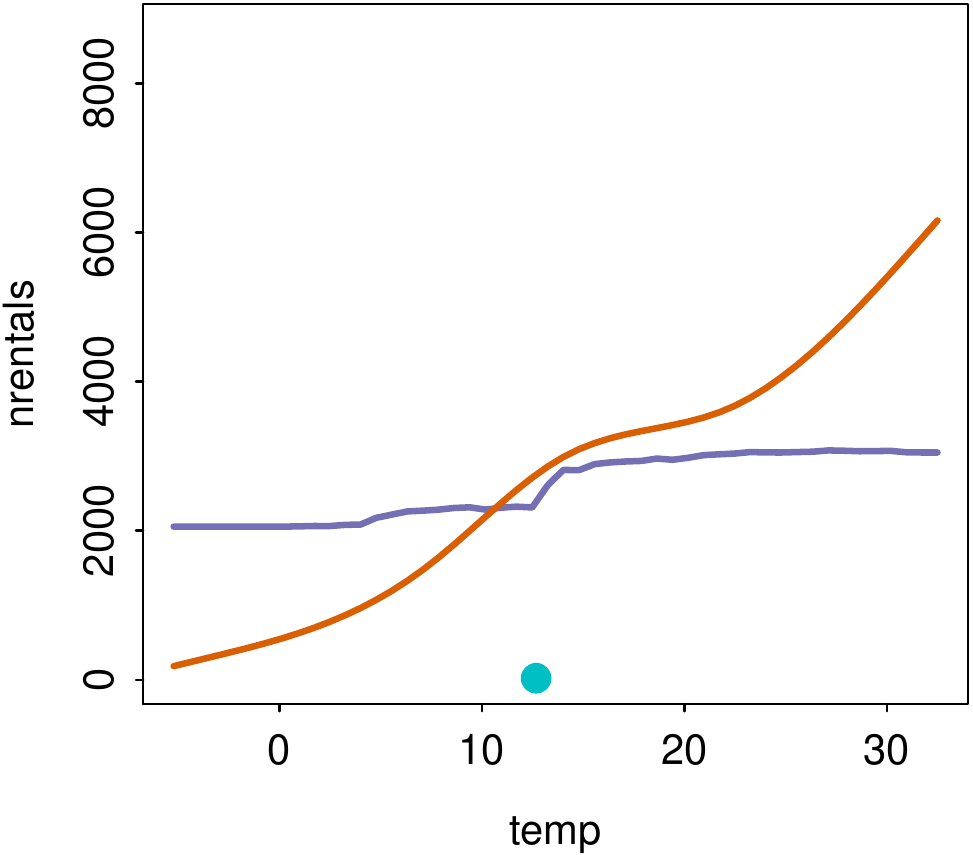} 
\begin{center}
\includegraphics[scale=.5,trim={0 80 0 80 },clip]{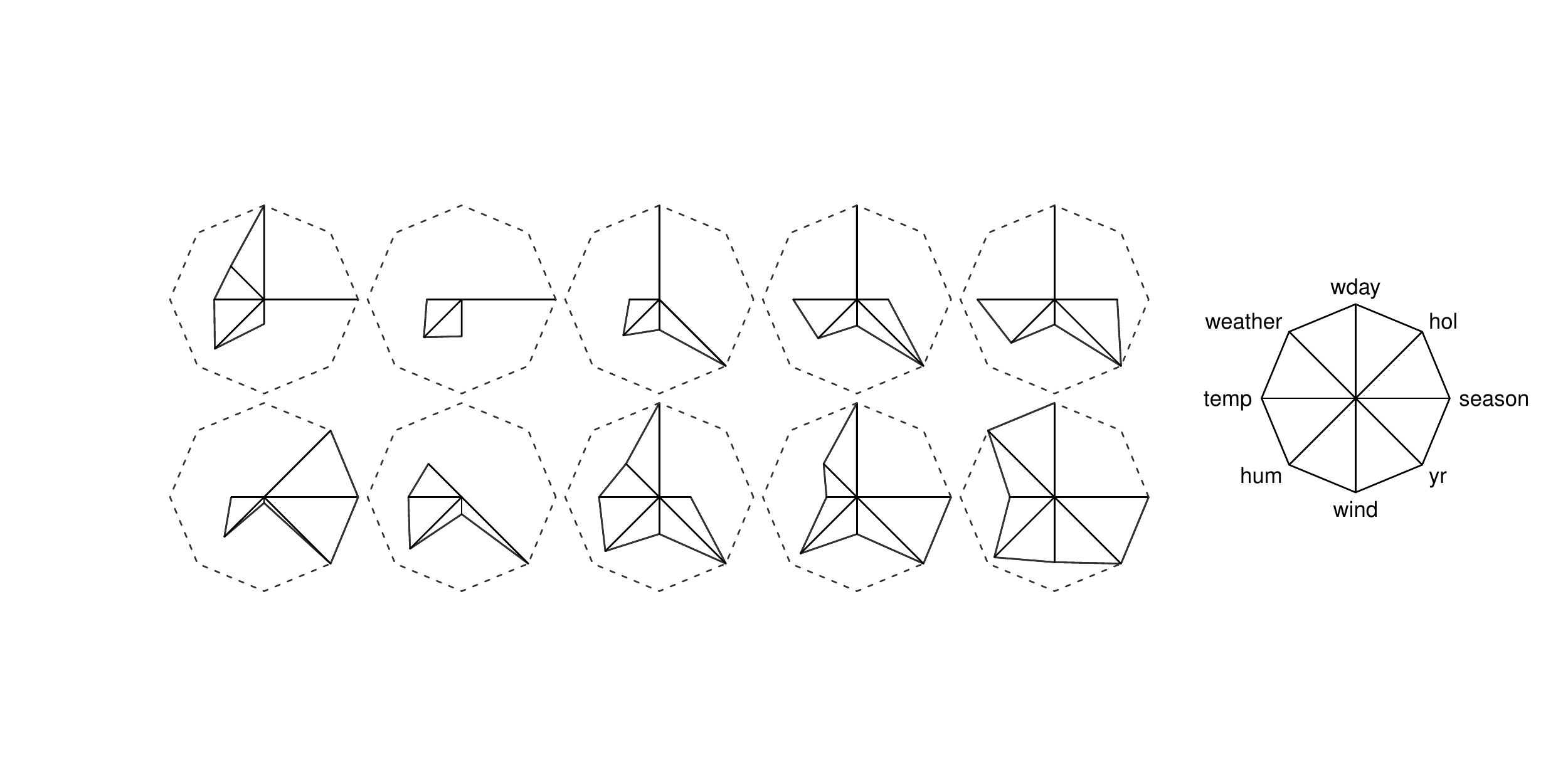} 
\end{center}
\caption{Tours of the bike data, nrentals versus temperature. Random forest fit in blue and gam in red, train and test observations  in light blue and pink respectively,
K-medoid tour in the first row,  lack of fit tour in second,  stars in rows 3,4 specify corresponding slices visited.
K-medoid shows gam fits better. Lack of fit tour stars show lack of fit occurs in 2012.}
\label{bikerfgam}
\end{figure}

For a deep-dive comparison of the two fits, we use the  tours of Section 2.2 to move through various slices, here using
the combined training and testing datasets.
Figure  \ref{bikerfgam} 
shows a k-medoid tour in the first row and lack of fit tour in the second row, with temp as the section variable and the remaining features
forming the condition variables. (Here for purposes of illustration both tours are constructed to be of length 5). The last two rows of Figure \ref{bikerfgam} show the condition variable settings for each of the ten tour points
as stars, where a long (short) radial line-segment indicates a high (low) value for a condition variable.
To the naked eye
the gam fit looks to give better results for most of the locations visited by the k-medoid tour. Switching to the lack of fit tour,
we see that the poorly-fit observation
in each  of the second row panels in Figure \ref{bikerfgam} has a large residual for both  the random forest and the gam fits.
Furthermore,  the poorly-fit observations identified
were all recorded in 2012, as is evident from the stars in the last row.

\subsection{Classification: Glaucoma data}
Glaucoma is an eye disease caused by damage to the optic nerve, which can lead to blindness if left untreated.
In \cite{glaucoma}, the authors explored various machine learning fits relating the occurrence of glaucoma
to age and various other features measured on the eye. The provided dataset comes pre-split into a training set of size 399 and
a test set of size 100. Here we focus on a random forest and a C5.0 classification tree \citep{c5}  fit to the training data.
 The random forest classified all training observations perfectly, mis-classifying just two test set observations,
 whereas the tree misclassified 20 and 6 cases for the training and test data respectively.
In a clinical setting however, as the authors in  \cite{glaucoma} pointed out, the results from a   classification tree
are easier to understand and implement. 

We will use interactive explorations to reduce the interpretability deficit for the random forest, and
to check if the simpler tree provides an adequate fit by comparison with the random forest, despite its
inferior test set performance.

\begin{figure}[htp]
\begin{center}
\includegraphics[scale=.6,trim={0 0 0 0},clip]{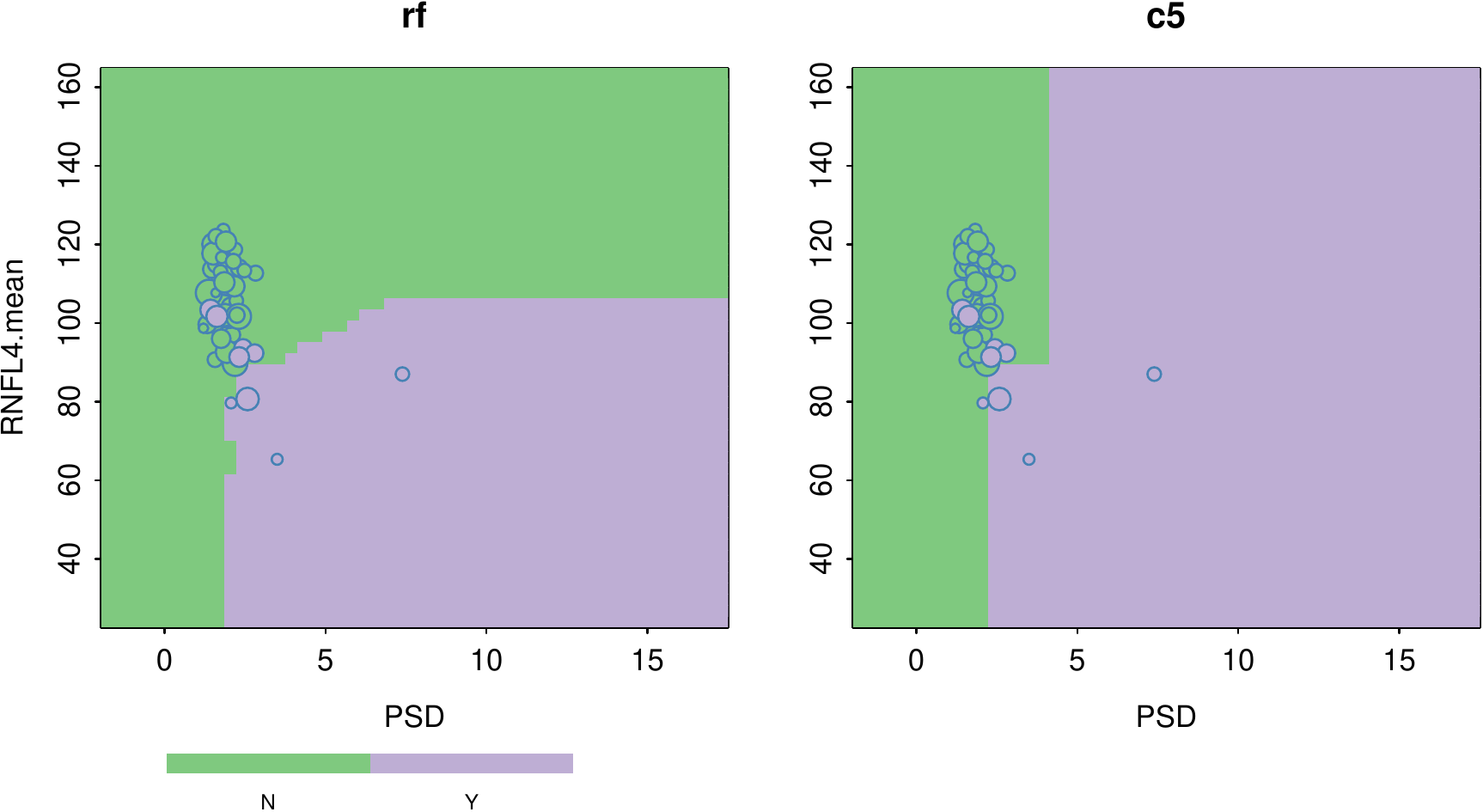} 
\end{center}
\caption{Section plots of random forest and tree fits for the glaucoma training data. Cases drawn in purple have glaucoma. In the region of these section plots with nearby observations, the fitted surfaces are the same.}
\label{glaucoma1}
\end{figure}

Figure  \ref{glaucoma1}
shows the training data with both classifiers. Here the section variables are  PSD and RNFL.mean (the two most important
features according to  random forest importance), 
and conditioning variables are set to values from the first case,
who is glaucoma free.  
Both classifiers give similar results for this condition, ignoring section plot regions with no data nearby.
Points whose color in the section plots disagrees with the background color of the classification surface
are not necessarily mis-classified, they are just near (according to the similarity score) a region whose
classification differs. Reducing the similarity threshold $\sigma$  to zero would show points whose values on the
conditioning predictors are identical to those of the first case, here just the first case itself, which is correctly
classified by both classifiers.
Clicking around on the condition selector plots and moving through the random, k-means and k-medoid tour paths shows that both classifiers give similar
classification surfaces for section predictors PSD and RNFL.mean, in areas where observations live.

\begin{figure}[htp]
\begin{center}
\begin{tabular}{cc}
\includegraphics[scale=.45,trim={0 0 0 0},clip]{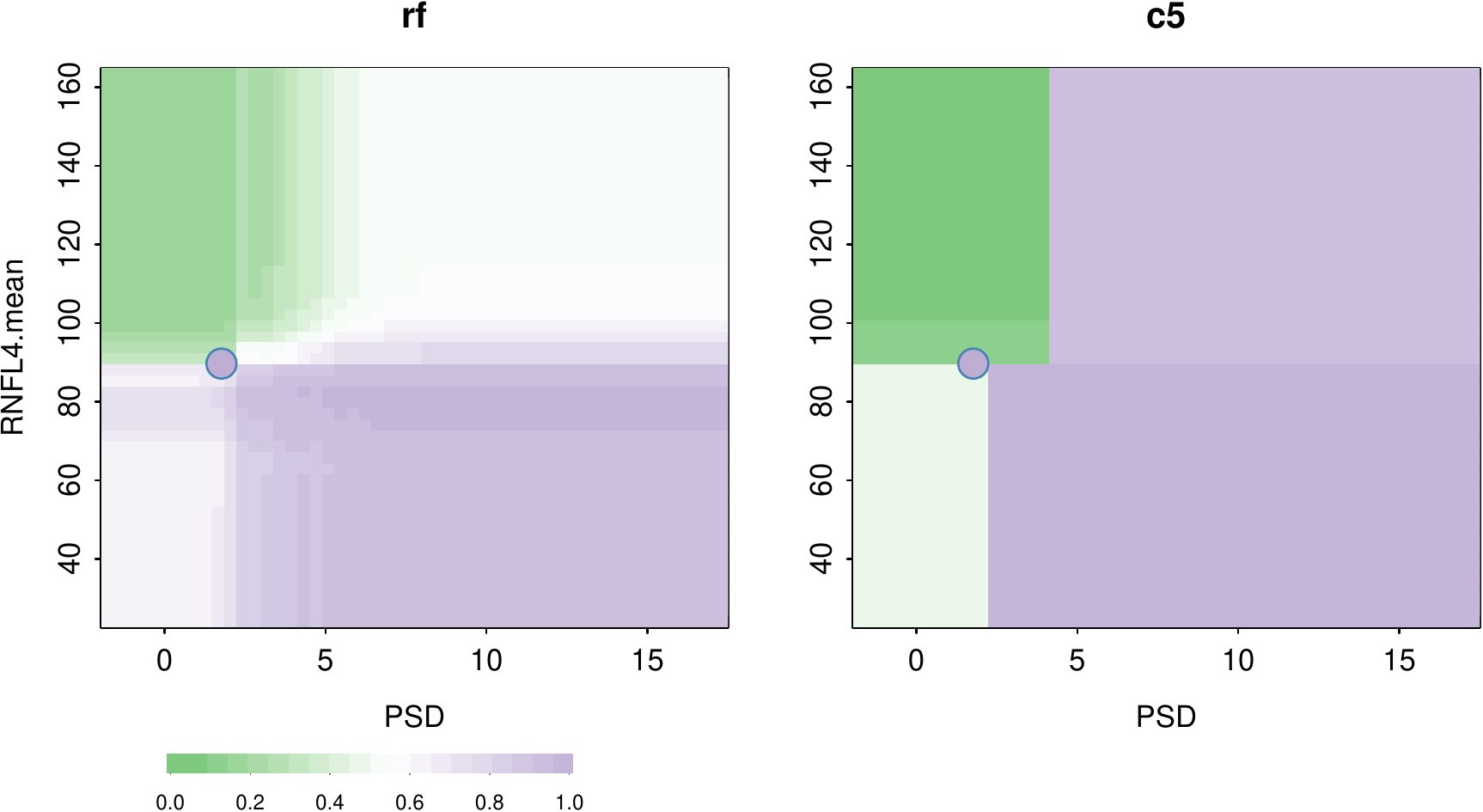}   & \includegraphics[scale=.45,trim={0 0 0 0},clip]{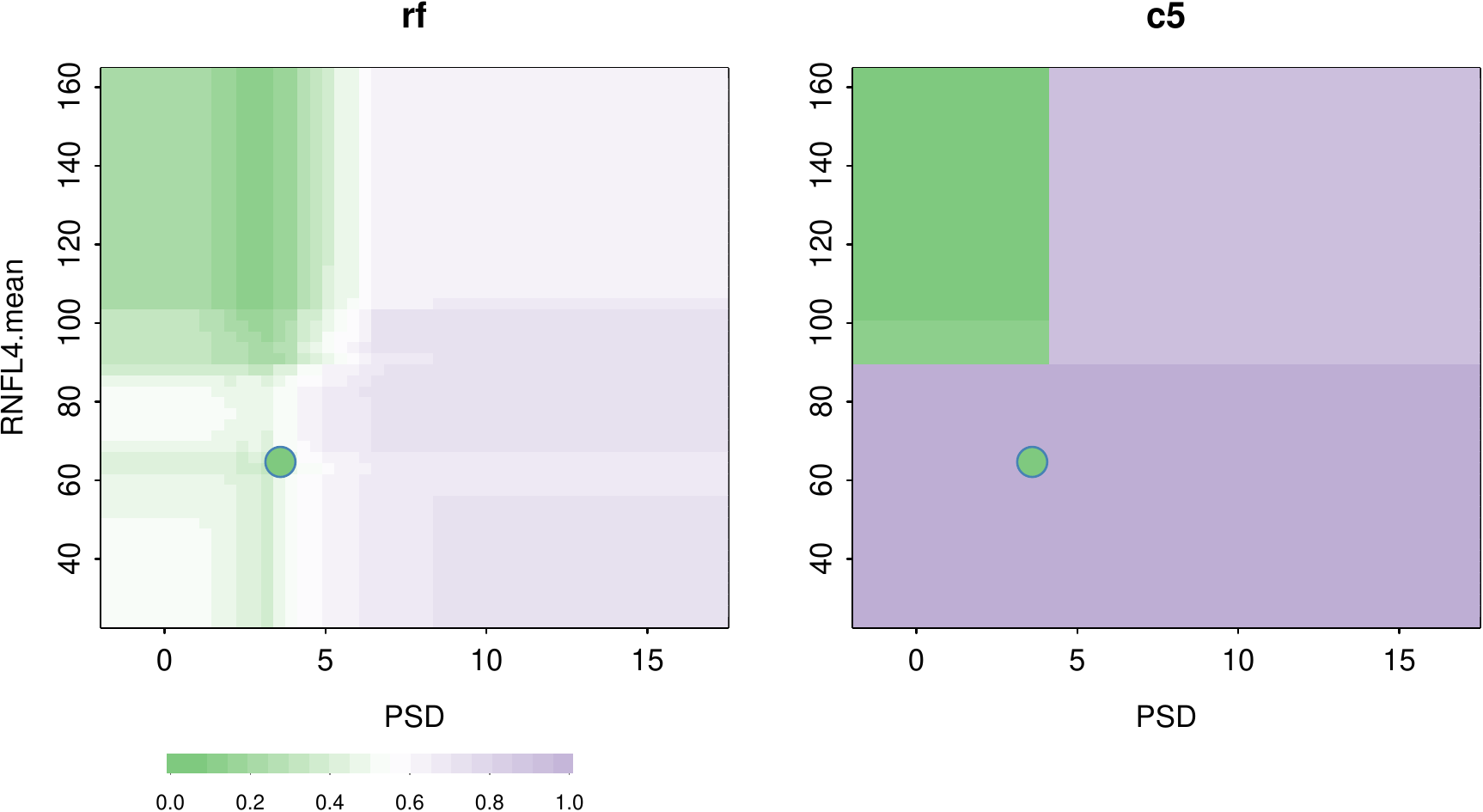}  \\
 (a) C5 false negative & (b)  C5 false positive
 \end{tabular}
 \end{center}
\caption{Glaucoma training data, random forest and tree fits, surface shows probability of glaucoma. Cases drawn in purple have glaucoma. Panels show cases wrongly classified by the tree, a false negative in (a) and false positive in (b).
}
\label{glaucoma2}
\end{figure}

Using the lack of fit tour to explore where the C5 tree gives incorrect predictions, in 
 Figure  \ref{glaucoma2}
the section plots show probability of glaucoma, on a green (for no glaucoma) to purple (for glaucoma) scale.
Here the similarity threshold $\sigma$ is set to zero, so only the
mis-classified observations are visible.
In the left hand side panel figure  \ref{glaucoma2}(a)
the C5 tree fit shows a false negative, which is quite close to the decision boundary. Though the random forest fit correctly classifies the
observation, it does not do so with high probability. Figure \ref{glaucoma2}(b)
shows a situation where the tree gives a false positive, which is well-removed from the decision boundary.
The random forest correctly predicts this observation as a negative, but the fitted surface is rough.
Generally training mis-classifications from the tree fit occur for PSD $\approx$ 2.5 and RNL4.mean $\approx$ 90,
where the random forest  probability surface is jumpy. So glaucoma prediction is this region is difficult 
based on this training dataset.

\subsection{Other application areas}

%
%

Typical ways to display clustering results include assigning colors to observations reflecting cluster membership,
and visualizing the colored observations in a scatterplot matrix, parallel coordinate plot or
in a plot of the first two principal components.
Some clustering algorithms such as k-means and model-based clustering algorithms offer predictions
for arbitrary points. The results of such algorithms can be visualized with our methodology. Section plots  show the
cluster assignment for various slices in the conditioning predictors.
As  in the classification example, we can compare clustering results, and check the cluster boundaries where
there is likely to be uncertainty in the cluster assignment. Suitable tours in this setting visit the centroid or medoid
of the data clusters. See the vignette   \url{https://cran.r-project.org/web/packages/condvis2/vignettes/mclust.html}  for an example.

One can also think of  density estimation algorithms as providing a ``fit''.
For such fits, the \code{CVpredict} function gives the density value, which
is renormalized over the section plot to integrate to 1. This way section plots show the density conditional
on the settings of the conditional variables.
With our condvis visualizations, we can compare two or more density functions or estimates by their
conditional densities for one or two section variables, assessing goodness of fit, and features such as number of modes and smoothness.
See the vignette   \url{https://cran.r-project.org/web/packages/condvis2/vignettes/mclust.html}  for an example.

\begin{figure}[htp]
\begin{center}
\includegraphics[scale=.3,trim={0 235 0 255},clip]{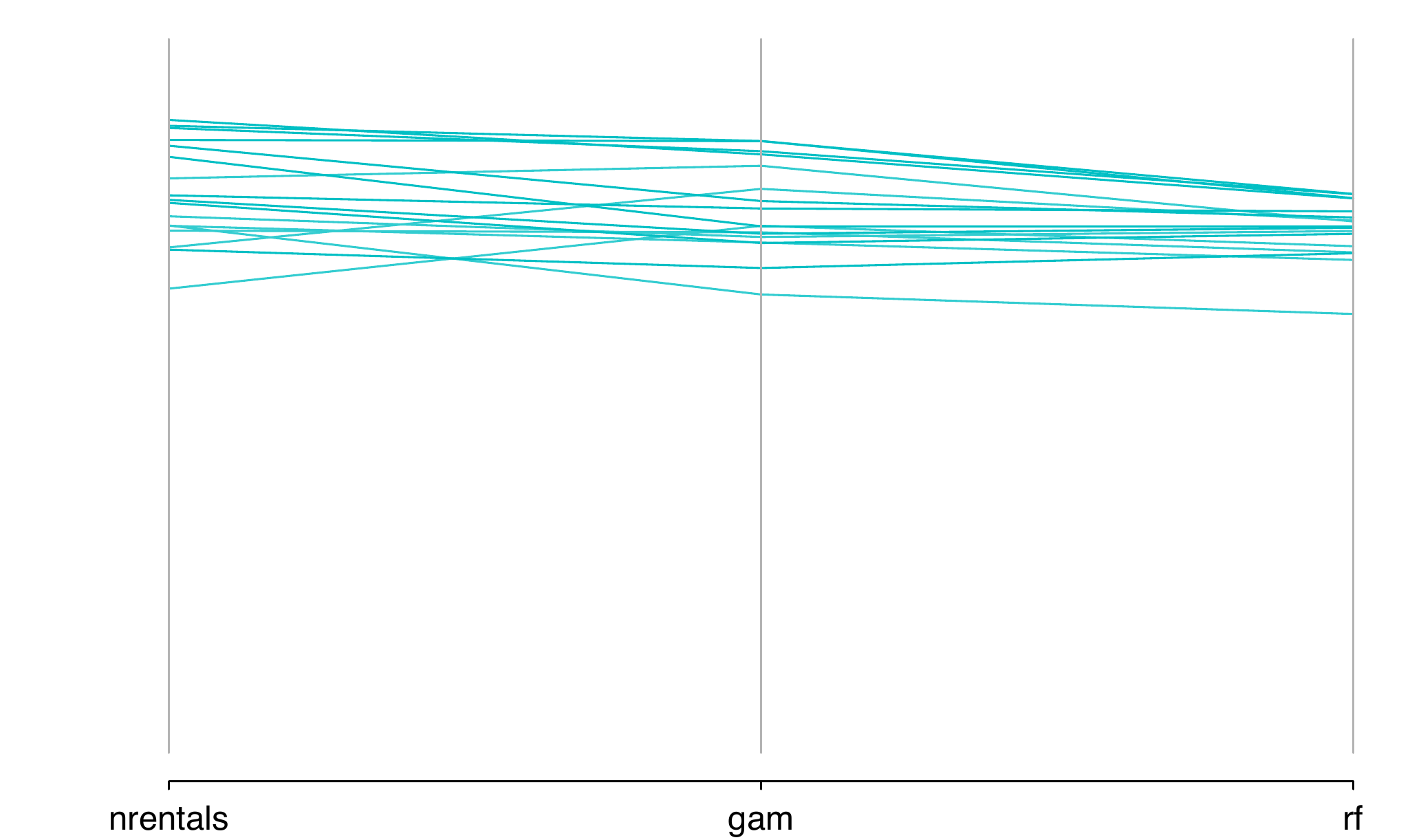}
\end{center}
\caption{Parallel coordinate plot showing response and predictions from gam and random forest  for  summer, weekday, good weather days in 2012 from the bike test data. For these
condition variables, the random forest underestimates nrentals by comparison with gam.}
\label{pcp}
\end{figure}

The ideas of conditional visualization may also be applied to situations where there is no fit function to be plotted.
In this case, the section plot shows observations for the section variables colored by similarity score which are determined to be
near the designated section point.
This is a situation where we provide section plots with $|\sect| > 2$.
One application of this is to compare predictions or residuals for an ensemble of model fits.
For the bike example of Section 4.1, consider the dataset augmented with predictions from the gam and random forest fits.
Figure \ref{pcp} 
shows a parallel coordinate of
three section variables, $y=$nrentals, $\hat{y}_{gam}$ and $\hat{y}_{rf}$ 
with the similarity threshold  set so that the plot shows all summer, weekday, good weather days in 2012.
The gam predictions are similar to the response as indicated by the mostly parallel line segment in the first panel, but the random forest underestimates the observed number of bike rentals.
This pattern  does not hold for 2011 data.

\section{Discussion}

We have described a new,  highly interactive application for deep-dive exploration of supervised and unsupervised learning model fits.
This  casts light on the black-box of machine learning algorithms, going far beyond simple numerical
summaries such as mean squared error, accuracy and predictor importance measures.
With interaction, the analyst can interrogate predictor effects and
pickup higher-order interactions in a way not possible with partial dependence and ICE plots,
explore goodness of fit to training or test datasets,
and compare multiple fits. Our new methodology will help machine learning practioners, educators
and students
seeking to interpret, understand and explain model results.
The application is currently useful for moderate sized datasets, up to 
100,000 cases and 30 predictors in our experience.
Beyond that, we recommend using case and predictor subsets to avoid lags in  response time which
make interactive use intolerable.

A previous paper \citep{condvis-jss} described  an early version of this project.
Since then, in \pkg{condvis2}  we have developed the project much further, 
and moved the implementation to  a Shiny platform which supports a far superior level of interactivity.
The choice of section plots and distance measures have been expanded. As an alternative to direct navigation through conditioning space,
we provide various algorithms for constructing tours, designed to visit non-empty slices 
(randomPath, kmeansPath and kmedPath) or slices showing lack of fit  (lofPath)
 or fit disparities (diffitsPath).
 We now offer an interface
to a wide and extensible range of machine learning fits, through  \code{CVpredict} methods, including
clustering algorithms and density fits. 
By providing an interface to the popular \pkg{caret}, \pkg{parsnip}, \pkg{mlr} and \pkg{mlr3} model-building platforms
our new interactive visualizations are widely accessible.

We recommend using variable importance measures to choose relevant section predictors, as in the case study of Section 4.2.
For pairs of variables, feature interaction measures such as the H-statistic \citep{friedman2008}
and its visualization available in \pkg{vivid} \citep{vivid} could be used to identify interesting pairs of section variables for interactive exploration.
New section touring methods could be developed to uncover other  plot patterns, but this needs to be
done in a computationally efficient way.
As mentioned previously, the tours presented here are quite different to  grand tours, as it is the slice that changes, not the projection.
In a recent paper \citep{tourr}, following on ideas from \cite{furnasbuja}, grand tours are combined with slicing, where slices are formed in the space
orthogonal to the current projection, but these techniques are not as yet  designed for the model fit setting.

There are some limitations in the specification of the section points through
interaction with the condition selector plots, beyond the fact that large numbers of predictors will not fit
in the space allocated to these plots (see Section 3.6).
If a factor has a large number of levels, then space becomes an issue. 
One possibility is to display only the most frequent categories in the condition selector plots,
gathering other categories into an ``other'' category, which of course is not selectable.
Also, we have not as yet addressed the situation where predictors are nested.

Currently we offer a choice of three distance measures (Euclidean, maxnorm and Gower) driving the similarity weights used in section plot displays.
Distances are calculated over predictors in $\cond$, other than the hidden predictors $F$.
Predictors are scaled to unit standard deviation  before distance is calculated which may not be appropriate for highly skewed predictors,
where a robust scaling is likely more suitable. We could also consider an option to
to interactively exclude some predictors from the distance calculation.

Other approaches could also be investigated for our section plot displays.
Currently, the section plot shows the fit
$\estf(\predictors_\sect = {\predictors_\sect}^g, \predictors_\cond = \predictorvals_\cond)$ versus ${\predictors_\sect}^g$,
overlaid on a subset of observations $(\predictors_{i\sect}, y_i)$, where $\predictors_{i\cond}$
belongs to the section around $\predictorvals_\cond$ (assuming $F = \emptyset$).
An alternative might be to display the average fit for observations in the section, that is
\[
{\rm ave}_{\predictors_{i\cond} \in {\rm sect(\predictorvals_\cond) }} \{ \estf(\predictors_\sect = {\predictors_\sect}^g, \predictors_\cond =\predictors_{i \cond}) \},
\]
or, form a weighted  average using the similarity weights.  Such a version of a section plot is analogous to a local version of a partial dependence plot.

We note that the popular lime algorithm of \cite{lime} also uses the concept of
conditioning to derive explanations for fits from machine learning models.  In their setup, all predictors are designated as  conditioning predictors, so $\sect = \emptyset$.
Lime explanations use a local ridge regression  to approximate   $\estf$ at $\predictors_\cond =\predictorvals$
 using  nearby sampled data, and the result is visualized in a barplot-type display of the local predictor contributions.
 For the purposes of the local approximation, the sampled data is  
 weighted by a similarity score. This contrasts with the approach presented here, where the similarity scores of Equation \ref{eq:simweight}
 are purely for visualization purposes. In \cite{wire}, we discussed how lime explanations could be generalized to the setting with
 one or two designated section variables, and this could usefully be embedded in an interactive application like ours.

\bibliographystyle{asa}

\bibliography{cvis}{}
\end{document}